\documentclass{article}
\usepackage[preprint]{log_2026}   %
\makeatletter\renewcommand{\@notice}{}\makeatother

\usepackage{amsmath, amssymb, amsthm}
\newtheorem{theorem}{Theorem}
\newtheorem{definition}{Definition}

\DeclareMathOperator{\tr}{tr}
\usepackage{booktabs}
\usepackage{array}
\usepackage{float}
\usepackage{placeins}
\usepackage{enumitem}
\usepackage{graphicx}
\usepackage{xcolor}
\usepackage{tikz}
\usetikzlibrary{arrows.meta, positioning, calc, backgrounds}
\usepackage[numbers,sort&compress]{natbib}

\usepackage{listings}
\definecolor{gracocmt}{RGB}{110,120,130}
\definecolor{gracokw}{RGB}{40,80,170}
\definecolor{gracostr}{RGB}{160,60,60}
\definecolor{gracobg}{RGB}{248,248,250}
\lstdefinestyle{graco}{
  language=Python,
  basicstyle=\ttfamily\small,
  keywordstyle=\color{gracokw},
  commentstyle=\color{gracocmt}\itshape,
  stringstyle=\color{gracostr},
  backgroundcolor=\color{gracobg},
  frame=single,
  rulecolor=\color{black!25},
  framesep=6pt,
  xleftmargin=6pt,
  showstringspaces=false,
  columns=fullflexible,
  keepspaces=true,
}

\newcommand{\Nnodes}{N}
\newcommand{\nclust}{n}
\newcommand{\Lift}{Q}          %
\newcommand{\Proj}{P}          %
\newcommand{\Prop}{S}          %
\newcommand{\Propc}{S_c}       %
\newcommand{\Sc}{\Propc}
\newcommand{\Adj}{A}
\newcommand{\Adjc}{A_c}
\newcommand{\Lap}{L}
\newcommand{\Lapc}{L_c}
\newcommand{\R}{\mathbb{R}}
\newcommand{\Ncal}{\mathcal{N}}

\title[GraCo: A Unified Toolkit of Graph Coarsening Algorithms]{GraCo: A Unified Toolkit of Graph Coarsening Algorithms and Metrics}

\author[A. Joly]{Antonin Joly \\ CNRS, IRISA, Rennes, France \\ \email{antonin.joly@irisa.fr}}

\begin{document}
\begin{center}
  \includegraphics[height=1.8cm]{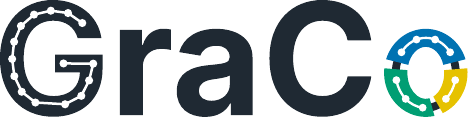}
\end{center}
\vspace{0.8em}
\maketitle

\begin{abstract}
Graph coarsening aims to reduce a graph to a smaller one that preserves chosen properties of the
original. It is increasingly used to scale graph learning to large graphs, as a non-differentiable step applied once before training.
Many coarsening methods have been proposed in recent years, each with its own implementation, notation and evaluation.
In this work, we introduce \textbf{GraCo}, a unified toolkit of graph coarsening methods.
It relies on a single definition: a coarsening is fixed by a partition of the nodes and by the Laplacian the coarsened graph is consistent with. The coarsened graph, and the reduction and lifting matrices that map signals from the original graph to the coarsened one and back, follow from these two choices.
We implement 17 methods from the literature and a random baseline, grouped into five families, with the same notation, inputs and outputs.
We also build a collection of metrics from different papers, applicable to any coarsening. 
We check the implementations on a synthetic stochastic block model, then
we report the runtime and peak memory of each method on Cora, PubMed, ogbn-arxiv and ogbn-products.
GraCo is open-source and available at \url{https://github.com/Antonin-J/graco}.
\end{abstract}

\section{Introduction}
\label{sec:intro}

Graph neural networks have become a standard tool for learning on graph-structured data,
but their cost grows with the size of the graph. On graphs with millions of nodes,
full-batch message passing exceeds the memory of a single device, and each epoch is slow.
Reducing the graph before learning is a common remedy, and it takes several forms: sampling a subgraph~\cite{Hu2013, hamilton2017inductive, chiang2019cluster}, sparsifying edges~\cite{spielman2011spectral, allen2015spectral, lee2018constructing}, condensing the graph into a small synthetic one~\cite{jin2022graph, zheng2024structure}, or coarsening it by grouping nodes into supernodes to form a coarsened graph~\cite{loukas2019graph, hashemi2024comprehensive}.
 \textbf{Graph coarsening} reduces a graph of $\Nnodes$ nodes to a smaller graph of $\nclust \ll \Nnodes$
supernodes that preserves chosen properties of the original. It is applied once, as a
preprocessing step, and is not required to be differentiable, which sets it apart from
graph pooling. 

Coarsening methods come from several communities: spectral graph
theory~\cite{loukas2019graph}, feature-aware
optimization~\cite{kumar2023featured, cohen2026spectral}, optimal
transport~\cite{xu2019scalable, ma2021unsupervised, taveras2026gromov}, and topology
preservation~\cite{wuscalable}. These methods differ in their input and output formats, their metrics and their notation. For instance, the reduction and the lifting matrices often swap names from one paper to the next, and each method is tied to one Laplacian. To our knowledge, there is no common interface through which a coarsening produced by one method can be scored by the metric of another, or reused outside its native setup.

This paper describes a \textbf{toolkit}: a single implementation of the
coarsening algorithms, documented here method by method, together with metrics that
apply to any coarsening.

\paragraph{Outline.}
Listing~\ref{lst:quickstart} shows the whole workflow, and the paper follows it in order:
\begin{enumerate}
\item A coarsening algorithm chooses a partition of the nodes into supernodes. Sec.~\ref{sec:design} presents this step and the 17 methods of the toolkit, grouped into five families.
\item The toolkit derives the products of the coarsening from this partition:
  \begin{itemize}
  \item the adjacency matrix of the coarsened graph (Sec.~\ref{sec:design});
  \item a \textbf{lifting matrix}, which maps a signal on the supernodes to the original nodes;
  \item a \textbf{reduction matrix}, which maps a signal of the original graph to the supernodes.
  \end{itemize}
  Sec.~\ref{sec:notation} explains the choice of the lifting, which amounts to choosing the Laplacian of the original graph that the coarsened graph is consistent with. By default, the reduction is the pseudoinverse of the lifting.
\item Metrics evaluate the quality of the coarsening (Sec.~\ref{sec:metrics}).
\end{enumerate}
Sec.~\ref{sec:sbm} checks the implementations on a synthetic graph drawn from a stochastic block model (SBM). Training graph neural networks on the coarsened graphs is planned for a later release of GraCo.

\begin{lstlisting}[style=graco,
  caption={Coarsening and scoring a graph with GraCo. Every method is called through the same
  \texttt{coarsen\_<name>} signature and returns only a partition;
  \texttt{build\_coarsening} derives the coarsened graph, the lifting $\Lift$, and the reduction
  $\Proj$. The same coarsening is then scored under metrics from three different papers.},
  label={lst:quickstart}, escapeinside={(*}{*)}]
from graco import get_dataset, coarsen_local_variation, build_coarsening
from graco.metrics import rsa, convolution_matching_error, max_acc

data = get_dataset("Cora")            # undirected, largest connected component

# every method has the same signature; change the suffix to change the algorithm
partition, info = coarsen_local_variation(data, r=0.5)

# the coarsened graph, the lifting Q and the reduction P then follow
co = build_coarsening(data, partition, laplacian_kind="normalized_self_loop")
co.coarsened                          # PyG Data with A_c and X_c
co.Q, co.P                            # lifting (N(*$\times$*)n), reduction (n(*$\times$*)N) matrices

# score the same coarsening under metrics from different papers
rsa(co, data)                         # spectral, Loukas 2019
convolution_matching_error(co, data)  # features, ConvMatch (Dickens 2024)
max_acc(co, data)                     # label-based upper bound
\end{lstlisting}

\section{A coarsening is a partition}
\label{sec:design}

A coarsening algorithm only chooses a partition of the nodes into supernodes, then the toolkit builds the coarsened graph from this partition. Any algorithm that assigns each node to a single supernode thus induces a coarsening, whether it was designed as graph clustering, community detection, or a proper coarsening. Methods differ only by the \textbf{criterion} they use to choose the partition. Every coarsening function takes the original graph and the coarsening ratio $r$, and its other parameters keep the defaults of its paper.

\paragraph{Setup.}
A graph $G$ on $\Nnodes$ nodes is described by a weighted adjacency matrix
$\Adj \in \R^{\Nnodes \times \Nnodes}$. Seven of the 17 methods also use a feature matrix $X \in \R^{\Nnodes \times d}$ (Table~\ref{tab:method-taxonomy}). Coarsening
reduces $G$ to a graph $G_c$ on $\nclust < \Nnodes$ supernodes, at \textbf{coarsening ratio}
$r = 1 - \nclust / \Nnodes$. A partition assigns each node to a single supernode. We write it as the matrix $\Lift_b \in \{0, 1\}^{\Nnodes \times \nclust}$, where
$(\Lift_b)_{ik} = 1$ means that node $i$ is mapped to supernode $k$.

\paragraph{Coarsened graph.}
The adjacency matrix of the coarsened graph is
\begin{equation}
  \Adjc = \Lift_b^\top \Adj\, \Lift_b,
  \label{eq:Ac}
\end{equation}
whose entry $(k, l)$ sums the weights of all edges between the two groups of original nodes. In
the toolkit this is \texttt{coarsened\_adjacency}$(\Adj, \Lift_b)$. By default the diagonal is
subtracted (flag \texttt{delete\_diagonal}), a convention inherited from Loukas~\cite{loukas2019graph}
that removes the heavily weighted self-loops which would otherwise perturb an iterative
edge-contraction scheme.

\paragraph{One interface for every algorithm.}
Every method of Sec.~\ref{sec:methods} has the same signature,
\texttt{coarsen\_<name>(data, r, **params)}, and returns a \texttt{Partition} together with
an \texttt{info} dictionary of run details (the achieved ratio, timing, method-specific
diagnostics). The achieved ratio differs from the requested ratio for several methods. METIS~\cite{karypis1997metis} enforces a minimum block size, so it cannot return many small supernodes, i.e.\ low ratios. It returns
$701$ supernodes where $1740$ were asked on Cora for $r=0.3$. The random baseline may leave some supernodes empty.
Graclus~\cite{dhillon2007weighted} matches each node with at most one neighbor per level, so its last level can overshoot the requested ratio. The dense-optimization methods can
collapse onto fewer supernodes than requested. The
\texttt{Partition} is the only algorithm-specific output; it is an assignment of the
$\Nnodes$ nodes to $\nclust$ supernodes.
Everything else comes from the toolkit, the same for all methods.

\paragraph{Input graph.}
The methods assume an undirected graph without self-loops. Several of them fail on a directed graph. The loader therefore symmetrizes each graph, strips self-loops and keeps the largest connected component, through three keyword arguments (\texttt{symmetrize}, \texttt{remove\_self\_loops}, \texttt{keep\_lcc}) that can be turned off. \texttt{preprocess\_graph} runs the same three steps on any PyG \texttt{Data} object. App.~\ref{app:datasets} gives the requirements of the methods and the effect of the three steps on our datasets.

\paragraph{Saving a coarsening.}
Saving the partition is enough: the toolkit rebuilds the rest of the coarsening on load (App.~\ref{app:storage}).

\subsection{The 17 methods}
\label{sec:methods}

\begin{table}[H]
\centering
\small
\caption{The 17 coarsening methods and the random baseline, grouped by family. \emph{Criterion} is the quantity each method optimizes or preserves. \emph{Impl.} is the fastest supported backend. \emph{Feat.} says whether the method uses the node features $X$ of the original graph. \emph{Scale} is the largest tier the method reaches within budget.}
\label{tab:method-taxonomy}
\resizebox{\linewidth}{!}{%
\begin{tabular}{llcclc}
\toprule
Method & Year & Criterion & Feat. & Impl. & Scale \\
\midrule
\multicolumn{6}{l}{\emph{Global clustering}} \\
Random                                 & --   & none                                & $\times$     & GPU     & very large    \\
METIS~\cite{karypis1997metis}          & 1997 & balanced edge cut                   & $\times$     & CPU     & very large    \\
Graclus~\cite{dhillon2007weighted}     & 2007 & normalized cut                      & $\times$     & GPU     & very large    \\
Leiden~\cite{traag2019louvain}         & 2019 & modularity                          & $\times$     & CPU     & very large    \\
UGC~\cite{kataria2024ugc}              & 2024 & feature-structure similarity        & $\checkmark$ & GPU     & very large    \\
GW-KGPC~\cite{taveras2026gromov}       & 2026 & Gromov-Wasserstein (GW) distortion  & $\times$     & GPU     & medium        \\
\midrule
\multicolumn{6}{l}{\emph{Representative selection}} \\
Kcenter~\cite{gonzalez1985clustering}  & 1985 & $k$-center covering                 & $\checkmark$ & GPU     & very large    \\
Herding~\cite{welling2009herding}      & 2009 & empirical-mean matching             & $\checkmark$ & GPU     & very large    \\
OTCoarsening~\cite{ma2021unsupervised} & 2021 & learned Wasserstein feature loss    & $\checkmark$ & GPU     & small         \\
\midrule
\multicolumn{6}{l}{\emph{Greedy contraction}} \\
Local Variation~\cite{loukas2019graph} & 2019 & restricted spectral approximation (RSA) & $\times$     & CPU     & very large    \\
ConvMatch~\cite{dickens2024graph}      & 2024 & convolution output                  & $\checkmark$ & GPU     & very large    \\
NOPE~\cite{bai2026rethinking2}          & 2026 & neighborhood interference           & $\checkmark$ & CPU     & very large$^{a}$ \\
GW-GPC~\cite{taveras2026gromov}        & 2026 & GW distortion                       & $\times$     & GPU     & small         \\
STPGC~\cite{wuscalable}                & 2026 & topology preservation               & $\times$     & CPU     & very large    \\
\midrule
\multicolumn{6}{l}{\emph{Recursive splitting}} \\
SGWL~\cite{xu2019scalable}             & 2019 & GW distance to a block target       & $\times$     & GPU     & large         \\
GBGC~\cite{xia2025gbgc, wang2026efficient} & 2025 & granular-ball quality            & $\times$     & CPU     & medium/very large \\
\midrule
\multicolumn{6}{l}{\emph{Dense optimization}} \\
FGC~\cite{kumar2023featured}           & 2023 & feature reconstruction + smoothness & $\checkmark$ & GPU     & medium$^{b}$        \\
IPE~\cite{cohen2026spectral}           & 2025 & spectral inner-product (Grassmann)  & $\times^{c}$ & GPU     & small         \\
\bottomrule
\end{tabular}}

\vspace{3pt}
{\footnotesize\raggedright
$^{a}$~NOPE reaches ogbn-products only through its NOPE$^{*}$ variant, in 15.5~h, beyond the twelve-hour budget.
$^{b}$~FGC reaches the large tier only at a high coarsening ratio, where
$\nclust$ is small enough for the dense assignment matrix to fit. It coarsens ogbn-arxiv at
$r = 0.99$ in 454~s (Table~\ref{tab:runtime}) and runs out of memory on ogbn-products.
$^{c}$~IPE uses $X$ only in its INGC variant. The default, SINGC,
minimizes a Grassmann term, a group-sparsity term and a log-determinant.\par}
\end{table}

We group the 17 methods of the toolkit and its random baseline into five families by how they assign the nodes to supernodes. For each method we give its principle, its source, and its \textbf{key hyperparameter}. By key hyperparameter we mean the parameter we would change first to adapt a method to a new graph: the one its paper names as its main hyperparameter, or the one that moved the result most in our runs. App.~\ref{app:hp-sensitivity} studies the hyperparameters of four methods in more depth, on the SBM. The default configuration of each method follows its paper, or its released code where the two differ. App.~\ref{app:method-catalogue} gives the detail and the complexity of each method, and App.~\ref{app:pseudocodes} lists these differences, together with our extensions.
Table~\ref{tab:method-taxonomy} also gives the criterion of each method, i.e.\ the quantity it optimizes or preserves. The same criterion can appear in several families. A Gromov-Wasserstein (GW) cost, for example, drives methods of three of them.

\paragraph{Global clustering.} These methods assign all nodes in one pass, by a cut, a modularity, a hash, a $k$-means, or at random. METIS also merges nodes level by level, but it scores the final partition, while greedy contraction scores each merge.
\begin{itemize}[leftmargin=*, itemsep=1pt]
\item \textbf{Random}, the baseline.
  \begin{itemize}[leftmargin=1.2em, nosep]
  \item Puts each node in a supernode drawn uniformly at random.
  \item Key hyperparameter: none. The variant \texttt{force\_equal\_size} gives all supernodes the same size, up to one node.
  \end{itemize}
\item \textbf{METIS}~\cite{karypis1997metis}.
  \begin{itemize}[leftmargin=1.2em, nosep]
  \item Cuts the graph into parts of balanced size, with as few edges between parts as possible.
  \item George Karypis and Vipin Kumar, METIS manual, University of Minnesota, 1997.
  \item Key hyperparameter: none.
  \end{itemize}
\item \textbf{Graclus}~\cite{dhillon2007weighted}.
  \begin{itemize}[leftmargin=1.2em, nosep]
  \item Minimizes the normalized cut by matching each node with a neighbor, level after level.
  \item Inderjit S. Dhillon, Yuqiang Guan and Brian Kulis, IEEE TPAMI, 2007.
  \item Key hyperparameter: none.
  \end{itemize}
\item \textbf{Leiden}~\cite{traag2019louvain}.
  \begin{itemize}[leftmargin=1.2em, nosep]
  \item Finds communities, i.e.\ groups of nodes with many edges inside and few to the rest, each of them connected.
  \item Vincent A. Traag, Ludo Waltman and Nees Jan van Eck, Scientific Reports, 2019.
  \item Key hyperparameter: \texttt{resolution}. A higher resolution gives more and smaller communities, and the toolkit searches it by default to reach the requested number of supernodes.
  \end{itemize}
\item \textbf{UGC}~\cite{kataria2024ugc}.
  \begin{itemize}[leftmargin=1.2em, nosep]
  \item Hashes each node from its features and its row of the adjacency matrix. The nodes that fall in the same bin form a supernode.
  \item Mohit Kataria, Sandeep Kumar and Jayadeva, NeurIPS, 2024.
  \item Key hyperparameter: \texttt{alpha}, the weight of the adjacency against the features in the hash. Raise it to give more weight to the structure. Its default is computed from the labels.
  \end{itemize}
\item \textbf{GW-KGPC}~\cite{taveras2026gromov}.
  \begin{itemize}[leftmargin=1.2em, nosep]
  \item Groups the nodes by a $k$-means on how much merging each pair of nodes would distort the graph, measured by a GW cost.
  \item Carlos A. Taveras, Santiago Segarra and C{\'e}sar A. Uribe, ICASSP, 2026.
  \item Key hyperparameter: \texttt{S\_kind}, the matrix whose distortion is measured, the adjacency by default.
  \end{itemize}
\end{itemize}

\paragraph{Representative selection.} These methods select $\nclust$ representative nodes and attach every other node to one of them.
\begin{itemize}[leftmargin=*, itemsep=1pt]
\item \textbf{Kcenter}~\cite{gonzalez1985clustering}.
  \begin{itemize}[leftmargin=1.2em, nosep]
  \item Smooths the features over the graph, then selects greedily the node with the smoothed features farthest from those of the nodes already selected. Every other node joins the closest selected node.
  \item Teofilo F. Gonzalez, Theoretical Computer Science, 1985.
  \item Key hyperparameter: \texttt{sgc\_hops}, the number of graph convolutions that smooth the features. Raising it smooths the features further over the graph, and $0$ uses the raw features, without using the graph structure.
  \end{itemize}
\item \textbf{Herding}~\cite{welling2009herding}.
  \begin{itemize}[leftmargin=1.2em, nosep]
  \item Smooths the features as Kcenter, then selects greedily the node that brings the mean smoothed features of the selected nodes closest to the mean over all nodes. The other nodes join as in Kcenter.
  \item Max Welling, ICML, 2009.
  \item Key hyperparameter: \texttt{sgc\_hops}, as for Kcenter.
  \end{itemize}
\item \textbf{OTCoarsening}~\cite{ma2021unsupervised}.
  \begin{itemize}[leftmargin=1.2em, nosep]
  \item A small graph neural network learns a score per node, and the best-scored nodes become the supernodes. Every other node joins the selected neighbor of highest score.
  \item Tengfei Ma and Jie Chen, AAAI, 2021.
  \item Key hyperparameters: \texttt{n\_epochs} and \texttt{lr}, the training of the scores. The defaults, $100$ epochs at a learning rate of $10^{-3}$, are enough on Cora and CiteSeer. A bigger graph might need more epochs.
  \end{itemize}
\end{itemize}

\paragraph{Greedy contraction.} These methods merge nodes level by level. At each level, they score candidate pairs or sets of nodes and merge the best ones, until the target ratio is reached.
\begin{itemize}[leftmargin=*, itemsep=1pt]
\item \textbf{Local variation}~\cite{loukas2019graph}, on edges or on whole neighborhoods.
  \begin{itemize}[leftmargin=1.2em, nosep]
  \item Merges the edges, or neighborhoods, whose contraction least changes the first $K$ eigenvectors of the Laplacian, i.e.\ the smoothest signals on the graph.
  \item Andreas Loukas, JMLR, 2019.
  \item Key hyperparameter: $K$, the number of eigenvectors to preserve, $100$ by default. On a very small graph, $100$ eigenvectors cover a large part of the spectrum, noise included, and a smaller $K$ may suit it better. One might also change \texttt{laplacian\_kind}, the Laplacian these eigenvectors come from.
  \end{itemize}
\item \textbf{ConvMatch}~\cite{dickens2024graph}.
  \begin{itemize}[leftmargin=1.2em, nosep]
  \item Merges the pairs of nodes that least change the output of one graph convolution when merged. The candidate pairs are nodes with close smoothed features.
  \item Charles Dickens, Edward Huang, Aishwarya Reganti, Jiong Zhu, Karthik Subbian and Danai Koutra, The Web Conference (Companion), 2024.
  \item Key hyperparameter: \texttt{merge\_candidates}, the pairs the method may merge. By default, as in the paper, these are pairs of nodes with similar features after one graph convolution. \texttt{knn+edges} adds the edges of the graph, so that more merges follow the graph structure.
  \end{itemize}
\item \textbf{NOPE}~\cite{bai2026rethinking2}.
  \begin{itemize}[leftmargin=1.2em, nosep]
  \item Merges the pair that least disturbs the neighbors the two nodes share.
  \item Xu Bai, Bin Lu, Kun Zhang, Shengbo Chen, Xinbing Wang, Chenghu Zhou and Meng Jin, ICML, 2026.
  \item Key hyperparameter: \texttt{variant}. NOPE$^{*}$ approximates the score of NOPE in closed form. The authors recommend NOPE at low to moderate ratios, and NOPE$^{*}$ when scalability matters more.
  \end{itemize}
\item \textbf{GW-GPC}~\cite{taveras2026gromov}.
  \begin{itemize}[leftmargin=1.2em, nosep]
  \item Merges, one pair at a time, the pair that least distorts the graph under the GW cost.
  \item Carlos A. Taveras, Santiago Segarra and C{\'e}sar A. Uribe, ICASSP, 2026.
  \item Key hyperparameter: \texttt{S\_kind}, as for GW-KGPC.
  \end{itemize}
\item \textbf{STPGC}~\cite{wuscalable}.
  \begin{itemize}[leftmargin=1.2em, nosep]
  \item Merges a node into a neighbor whose neighborhood contains its own, which preserves the topology of the graph, and relaxes this rule when no such pair is left.
  \item Xiang Wu, Rong-Hua Li, Xunkai Li, Kangfei Zhao, Hongchao Qin and Guoren Wang, ICML, 2026.
  \item Key hyperparameter: $\theta_2$ (\texttt{theta2\_ratio}), $1\%$ of $\Nnodes$ by default, which sets how soon the merge rule is relaxed once no exact merge is left. A larger $\theta_2$ relaxes it sooner, which may suit graphs where few neighborhoods are nested.
  \end{itemize}
\end{itemize}

\paragraph{Recursive splitting.} These methods first cut the graph into a few large parts, then cut each part again to refine the partition.
\begin{itemize}[leftmargin=*, itemsep=1pt]
\item \textbf{SGWL}~\cite{xu2019scalable}.
  \begin{itemize}[leftmargin=1.2em, nosep]
  \item Splits the graph in two by an optimal transport problem with a GW cost, then splits the largest part again until $\nclust$ parts exist.
  \item Hongteng Xu, Dixin Luo and Lawrence Carin, NeurIPS, 2019.
  \item Key hyperparameter: \texttt{beta}, the weight of the regularization of the transport problem, which has to match the scale of the cost.
  \end{itemize}
\item \textbf{GBGC}~\cite{xia2025gbgc, wang2026efficient}.
  \begin{itemize}[leftmargin=1.2em, nosep]
  \item Cuts the graph into about $\sqrt{\Nnodes}$ balls grown from the nodes of highest degree, then splits a ball in two whenever the split raises a quality score. It stops when no split helps, so the ratio it reaches depends on the graph.
  \item Shuyin Xia, Guan Wang, Gaojie Xu, Sen Zhao and Guoyin Wang, IJCAI, 2025. Follow-up work by Guan Wang, Shuyin Xia, Lei Qian, Tao Wu, Guoyin Wang, Yi Wang and Wei Wang, preprint, 2026.
  \item Key hyperparameter: \texttt{init}. The METIS start of the follow-up work~\cite{wang2026efficient} gives fewer initial balls, so the method can reach higher coarsening ratios.
  \end{itemize}
\end{itemize}

\paragraph{Dense optimization.} These methods learn a soft assignment of the nodes to the supernodes, i.e.\ a dense $\Nnodes \times \nclust$ matrix. They train it by gradient descent on a differentiable loss. They then round it to a partition by a row-wise argmax. A fundamental limitation of the family is the size of this assignment matrix: $\mathcal{O}(\Nnodes \nclust)$ memory, and $\mathcal{O}(\Nnodes \nclust^2)$ time per iteration. It fits on a commodity GPU only at small $\Nnodes$, or at high ratios where $\nclust$ is small. These methods also have many hyperparameters: several loss weights, a learning rate and an iteration count. They were the hardest methods to tune in our runs (App.~\ref{app:hp-sensitivity}). Below, we name for each method the hyperparameters that moved its result most.
\begin{itemize}[leftmargin=*, itemsep=1pt]
\item \textbf{FGC}~\cite{kumar2023featured}.
  \begin{itemize}[leftmargin=1.2em, nosep]
  \item Learns an assignment that preserves both the structure and the features of the graph.
  \item Manoj Kumar, Anurag Sharma, Shashwat Saxena and Sandeep Kumar, ICML, 2023.
  \item Key hyperparameters: \texttt{lr}, then \texttt{alpha} and \texttt{lam}. A small graph might need a larger step than the default $10^{-5}$. Lowering \texttt{alpha}, the weight of the feature reconstruction, gives more weight to the structure. Too high a \texttt{lam}, the weight of a group penalty on the assignment, leaves supernodes empty.
  \end{itemize}
\item \textbf{IPE}~\cite{cohen2026spectral}.
  \begin{itemize}[leftmargin=1.2em, nosep]
  \item Same principle as FGC, with a spectral objective in place of the reconstruction terms.
  \item Ido Cohen and Ronen Talmon, NeurIPS, 2025.
  \item Key hyperparameters: \texttt{lr}, then \texttt{k\_grassmann}. The default step, $10^{-4}$, comes from the runs of the paper on Cora, and a smaller graph might need a larger one. \texttt{k\_grassmann} is the number of leading eigenvectors of the Laplacian the coarsened graph is aligned with. Close to the number of classes, it pulls the partition toward the classes, but it may then return fewer supernodes than requested.
  \end{itemize}
\end{itemize}

Node-elimination reductions such as Kron's Schur complement~\cite{dorfler2012kron}, which
shrink a graph without partitioning it into supernodes, fall outside this partition-based
view and are out of scope. Data-driven refinements that re-weight an existing coarsened graph,
such as GOREN~\cite{cai2021graph}, are also out of scope.
Topological spatial coarsening~\cite{calissano2025topological} does fit the partition view, but it contracts edges by their length, which requires node positions in $\R^d$ that the graphs considered here do not have.

Table~\ref{tab:method-taxonomy} and the summaries of App.~\ref{app:method-catalogue} give each method a \emph{Scalability} label, the largest of four tiers it reaches, namely \textbf{small} (a few thousand nodes), \textbf{medium} (tens of thousands), \textbf{large} (hundreds of thousands, the scale of ogbn-arxiv), and \textbf{very large} (millions, the scale of ogbn-products). We call a method scalable at a tier when our implementation coarsens a graph of that size on a single A40 GPU of 48\,GB within the budget of the tier, which is one hour up to medium, three hours for large, and twelve hours for very large. We filled the labels from the runs of Table~\ref{tab:runtime}.

\subsection{Runtime and memory}
\label{sec:runtime}

\begin{table}[t]
\centering
\footnotesize
\setlength{\tabcolsep}{3pt}
\newcommand{\nomeas}{\textcolor{black!35}{$\times$}}
\caption{Runtime (s) and peak memory (MB) of each method over the four tiers, at $r = 0.9$ on Cora and PubMed, $0.99$ on ogbn-arxiv and $0.999$ on ogbn-products. Memory is the peak above the floor described in Sec.~\ref{sec:runtime}. A dash marks a method that never allocates on the GPU, and a grey cross an entry the run did not reach.}
\label{tab:runtime}
\begin{tabular}{lccccccccccccc}
\toprule
& \multicolumn{3}{c}{Cora} & \multicolumn{3}{c}{PubMed} & \multicolumn{3}{c}{ogbn-arxiv} & \multicolumn{3}{c}{ogbn-products} \\
\cmidrule(lr){2-4}\cmidrule(lr){5-7}\cmidrule(lr){8-10}\cmidrule(lr){11-13}
Method & time & CPU & GPU & time & CPU & GPU & time & CPU & GPU & time & CPU & GPU \\
\midrule
Random         & $<$0.01 & 0 & -- & $<$0.01 & 11 & -- & 0.01 & 0 & -- & 0.12 & 0 & -- \\
METIS          & 0.07 & 0 & -- & 0.83 & 16 & -- & 4.5 & 163 & -- & 84.6 & 0 & -- \\
Graclus        & 0.04 & 3 & -- & 0.21 & 23 & -- & 1.5 & 152 & -- & 49.7 & 0 & -- \\
Leiden$^{\ddagger}$ & 2.9 & 32 & -- & 34.7 & 64 & -- & 2718 & 355 & -- & 33523 & 0 & -- \\
UGC            & 0.25 & 56 & -- & 0.54 & 216 & -- & 3.4 & 1496 & -- & 35.1 & 4628 & -- \\
GW-KGPC        & 0.63 & 565 & 392 & 51.2 & 1587 & 23753 & \nomeas & OOM & \nomeas & \nomeas & OOM & \nomeas \\
Kcenter        & 0.55 & 466 & 66 & 1.8 & 455 & 279 & 3.0 & 310 & 481 & 44.9 & 0 & 11132 \\
Herding        & 0.51 & 444 & 66 & 1.7 & 423 & 279 & 2.6 & 282 & 481 & 40.6 & 0 & 11132 \\
OTCoarsening   & 11.6 & 579 & 1042 & \nomeas & OOM & \nomeas & \nomeas & OOM & \nomeas & \nomeas & OOM & \nomeas \\
Local Var.     & 0.98 & 101 & -- & 3.3 & 300 & -- & 100.0 & 2492 & -- & 3164 & 1979 & -- \\
Local Var. nbh. & 10.8 & 230 & -- & 10.4 & 302 & -- & 288 & 10130 & -- & 13772 & 8198 & -- \\
ConvMatch      & 4.1 & 588 & 295 & 2.9 & 691 & 815 & 19.7 & 1380 & 1452 & 2402 & 0 & 16558 \\
A-ConvMatch    & 5.2 & 555 & 215 & 2.4 & 634 & 616 & 18.7 & 1371 & 1443 & 2368 & 0 & 16571 \\
NOPE           & 0.80 & 16 & -- & 9.8 & 69 & -- & 1587 & 479 & -- & \nomeas & OOT & \nomeas \\
NOPE$^{*}$   & 0.21 & 14 & -- & 3.1 & 66 & -- & 183 & 425 & -- & 55936 & 7054 & -- \\
GW-GPC         & 6.4 & 527 & 416 & \nomeas & OOT & \nomeas & \nomeas & OOM & \nomeas & \nomeas & OOM & \nomeas \\
STPGC          & 0.14 & 1 & -- & 1.9 & 27 & -- & 106 & 265 & -- & 16666 & 1329 & -- \\
SGWL           & 6.3 & 478 & 9 & 61.5 & 486 & 16 & 1124 & 375 & 246 & \nomeas & OOT & \nomeas \\
GBGC$^{\dagger}$ & $<$0.01 & 2 & -- & 0.07 & 26 & -- & 2.3 & 159 & -- & 142 & 0 & -- \\
GBGC-METIS     & 0.07 & 2 & -- & 0.59 & 29 & -- & 7.3 & 370 & -- & 362 & 4561 & -- \\
FGC            & 6.3 & 433 & 80 & 94.6 & 576 & 1350 & 454 & 2117 & 9013 & \nomeas & OOM & \nomeas \\
IPE            & 9.9 & 462 & 31 & \nomeas & FAIL & \nomeas & \nomeas & OOT & \nomeas & \nomeas & OOM & \nomeas \\
\bottomrule
\end{tabular}

\vspace{3pt}
{\footnotesize\raggedright
$^{\dagger}$~GBGC stops on its own quality criterion and returns 401, 7109, 28254 and
305151 supernodes where the four tiers ask for 249, 1972, 1694 and 2386. Its time
therefore covers a different job. On ogbn-products it built a coarsened graph a hundred times
larger than every other row of that column. GBGC-METIS starts from the
$\lceil\sqrt{\Nnodes}\rceil$ parts METIS returns and splits them until the requested count is
reached, which it reaches on all four tiers.
$^{\ddagger}$~On ogbn-products, the resolution search of Leiden stops before a third run would overrun the twelve-hour budget. It returns the closest partition it met, $599$ supernodes for the $2386$ asked.\par}
\end{table}

Table~\ref{tab:runtime} reports the wall-clock time and peak memory of each method over
the four tiers, from Cora to ogbn-products. The ratio of each tier is chosen so that the
coarsened graph has between 250 and 2500 supernodes on all four graphs. Only the size of
the input then changes from one graph to the next. Every row was measured on the same CPU, an AMD
EPYC 7413, with an NVIDIA A40 of 48~GB, and torch pinned to 24 threads. 
Three marks replace a measurement. \textbf{OOM} stands
for an allocation the host or the device refused, \textbf{OOT} for an overrun of the
per-method budget, one hour on Cora and PubMed, three on ogbn-arxiv and twelve on
ogbn-products, and \textbf{FAIL} for a solver that did not converge. Both memory columns report the peak above the floor each method needs before it starts, namely torch and the loaded graph. This floor is 710, 730 and 1006\,MB on the first three graphs, and 23.6\,GB on ogbn-products. On ogbn-products, a method whose own peak stays below this floor thus has a $0$ in that column. A method still has to reach the requested size for its
time to mean anything. GBGC stops short of it on all tiers, and Leiden on ogbn-products.

Note however that the top tier is \emph{not} reached everywhere (Table~\ref{tab:runtime}). OTCoarsening,
GW-GPC and IPE remain capped at a few thousand nodes, and GW-KGPC at tens of thousands. OTCoarsening and GW-KGPC run out of memory on their dense matrices, GW-GPC runs out of time with its $\mathcal{O}(\Nnodes^4)$ search, and IPE does not converge. FGC passes the medium tier only at very high ratios (Table~\ref{tab:method-taxonomy}).

\section{Choosing the lifting and the reduction matrices}
\label{sec:notation}

Sec.~\ref{sec:design} builds a partition of the nodes into supernodes, and from it the coarsened graph. A coarsening also has to move signals between the two graphs. A signal on the nodes, such as a feature, lives in dimension $\Nnodes$ on the original graph and in dimension $\nclust$ on the coarsened one. Moving it from one graph to the other takes two matrices. The \textbf{reduction matrix} $\Proj \in \R^{\nclust \times \Nnodes}$ coarsens a signal of the original graph into a signal of the coarsened one. The \textbf{lifting matrix} $\Lift \in \R^{\Nnodes \times \nclust}$ lifts a coarse signal back to the original nodes. For $x \in \R^{\Nnodes}$, the coarsened signal $x_c \in \R^{\nclust}$ and the re-lifted signal $\tilde{x} \in \R^{\Nnodes}$ are
\begin{equation}
  x_c = \Proj x, \qquad \tilde{x} = \Lift x_c = \Pi x, \qquad \Pi = \Lift \Proj.
  \label{eq:signals}
\end{equation}
The same operators act row-wise on the feature matrix, $X_c = \Proj X$ and $\tilde{X} = \Lift \Proj X$.
Several of the metrics of Sec.~\ref{sec:metrics} are computed with these operators, either on the reconstructed signal $\tilde{x}$ or on the coarse operators that produce it.
We follow the setting of~\cite{joly2026taxonomy, joly2026core}, which builds on the graph
coarsening framework of Loukas~\cite{loukas2019graph}. In this setting, choosing the lifting amounts to choosing the Laplacian of the original graph that the coarsened graph is consistent with. This section presents this choice, then briefly the choice of the reduction. Neither choice changes the partition or the coarse adjacency $\Adjc$.

\paragraph{Binary lifting.}
The combinatorial Laplacian of $G$ is $\mathcal{L} = \mathcal{L}(\Adj) := D - \Adj$, with $D = \operatorname{diag}(\Adj \mathbf{1}_\Nnodes)$
the degree matrix. The partition matrix $\Lift_b$ of Sec.~\ref{sec:design} has exactly one non-zero coefficient per row, equal to $1$, since each node belongs to a single supernode.

\begin{definition}[Well-partitioned matrix]
\label{def:wp}
A matrix is \textbf{well-partitioned} if it has exactly one non-zero coefficient per row.
\end{definition}
In other words, a well-partitioned lifting maps each node of the original graph to a single supernode.

We call \textbf{lifting} of the partition any matrix $\Lift \in \R^{\Nnodes\times \nclust}$
with the non-zero pattern of $\Lift_b$, its coefficients may differ from $1$. Such a $\Lift$ is well-partitioned. A matrix is \textbf{binary} if all its coefficients are $0$ or $1$, and $\Lift_b$ is the binary lifting.
As Loukas~\cite{loukas2019graph} shows, up to a scale factor, the binary lifting is the only one for
which the product of Eq.~\eqref{eq:Ac} also gives the coarse combinatorial Laplacian.

\begin{theorem}[Loukas~\cite{loukas2019graph}]
\label{lem:loukas}
Let $\Lift$ be a lifting. The following two properties are equivalent: (i) $\Lift$ is
proportional to a binary matrix; (ii) for all $\Adj$,
$\mathcal{L}(\Lift^\top \Adj \Lift) = \Lift^\top \mathcal{L}(\Adj)\, \Lift$.
\end{theorem}

\paragraph{Normalized liftings.}
So far the lifting is binary, and the coarsened graph is consistent with the combinatorial Laplacian as defined by Loukas. For a graph neural network the object worth preserving is not necessarily the
combinatorial Laplacian, but it might be associated with the operator a graph convolution propagates with. We therefore adopt the
broader family of $\Delta$-Laplacians~\cite{joly2024graph, joly2026taxonomy},
\begin{equation}
  \Lap = \Lap(\Adj) := \Delta\, \mathcal{L}\, \Delta,
  \label{eq:delta-lap}
\end{equation}
where $\Delta = \Delta(\Adj)$ is a strictly positive diagonal matrix. Setting $\Delta = I_\Nnodes$
recovers the combinatorial Laplacian, $\Delta = D^{-1/2}$ the symmetric normalized one, and
$\Delta = (D + I_\Nnodes)^{-1/2}$ the self-loop normalized Laplacian, for which
$\Lap = I_\Nnodes - \Prop$ with
$\Prop = (D + I_\Nnodes)^{-1/2}(\Adj + I_\Nnodes)(D + I_\Nnodes)^{-1/2}$
the propagation matrix of the graph convolutional network of Kipf and Welling~\cite{kipf2017semisupervised}.
This operator $\Prop$ reappears in several metrics of Sec.~\ref{sec:metrics}. The toolkit provides these three choices through a
\texttt{laplacian\_kind} argument (\texttt{combinatorial}, \texttt{normalized},
\texttt{normalized\_self\_loop}). The $\Delta$-Laplacian family is more general, but these
three cover the current release.

Consistency with such a normalized Laplacian requires a degree-aware lifting,
\begin{equation}
  \Lift = \Delta^{-1}\, \Lift_b\, \Delta_c,
  \label{eq:gen-lift}
\end{equation}
where $\Delta_c = \Delta(\Adjc)$. This lifting of the partition is no longer binary.
Theorem~\ref{lem:loukas} extends to the whole family.

\begin{theorem}[Consistency, extension~\cite{joly2026taxonomy}]
\label{lem:consistency}
Let $\Lift'$ be a lifting, $\Adjc' = \Lift'^\top \Adj \Lift'$ and
$\Lift = \Delta^{-1} \Lift' \Delta(\Adjc')$. The following two properties are equivalent:
(i) $\Lift'$ is proportional to a binary matrix; (ii) for all $\Adj$,
$\Lap(\Adjc') = \Lift^\top \Lap(\Adj)\, \Lift$.
\end{theorem}

With $\Lift' = \Lift_b$, it gives the lifting of Eq.~\eqref{eq:gen-lift}. Theorem~\ref{lem:loukas}
is its combinatorial case $\Delta = I_\Nnodes$. In the toolkit the lifting of
Eq.~\eqref{eq:gen-lift} is \texttt{Q\_normalized\_from\_partition}, selected by a
\texttt{laplacian\_kind} argument. Its default is the self-loop normalized Laplacian, since a graph
convolution propagates with it. The binary lifting remains available.
\texttt{convert\_\allowbreak laplacian\_kind\_\allowbreak coarsening} rebuilds $\Lift$ and $\Proj$ under another \texttt{laplacian\_kind}.

\paragraph{Choosing the reduction.}
The reduction $\Proj$ enters neither $\Adjc$ nor the coarse Laplacian, only the
coarsened signal $x_c = \Proj x$ and the reconstruction $\Lift \Proj x$. A GNN trained on the coarsened graph thus uses the features $X_c = \Proj X$. By default, $\Proj$ is the pseudoinverse $\Proj = \Lift^{+}$,
in closed form on a well-partitioned lifting (\texttt{pinv\_well\_partitioned\_torch}).
For the binary lifting, it averages each feature over its supernode. It is the default of most of the literature. Recent work~\cite{joly2026taxonomy} showed that, once the partition and the lifting are fixed, $\Proj$ still has degrees of freedom that can be optimized. The module \texttt{graco.refinement} implements this optimization. We refer to~\cite{joly2026taxonomy} for more details.

\paragraph{From a partition to a coarsening.}
A single call, \texttt{build\_coarsening(data, partition, laplacian\_kind)}, builds the coarse adjacency $\Adjc$ of Eq.~\eqref{eq:Ac}, the lifting $\Lift$ for the chosen \texttt{laplacian\_kind}, the reduction $\Proj = \Lift^{+}$ and the coarse features $X_c = \Proj X$. It returns them in a \texttt{Coarsening}, together with the partition and the run metadata (method, requested and achieved ratios, \texttt{laplacian\_kind}). Its coarsened graph is a standalone PyG \texttt{Data} object, on which any message-passing model runs unchanged.
For the methods that choose their partition with a Laplacian, such as local variation, the \texttt{laplacian\_kind} of the lifting matrix is independent of the Laplacian the coarsening algorithm uses internally (App.~\ref{app:laplacian-table}).

\section{Coarsening metrics}
\label{sec:metrics}

Every metric of this section takes the \texttt{Coarsening} of Sec.~\ref{sec:notation}, and the original graph when it compares the two graphs, \texttt{metric(coarsening, original)}. A coarse-only metric takes the coarsening alone, so a large original graph can be freed from memory once the coarsening is built. Any metric can be used on any coarsening. A coarsening designed to minimize one criterion can thus be measured under the criterion of a competing method. App.~\ref{app:cross-scoring} scores each method on Cora and PubMed under four of these metrics.

Table~\ref{tab:metrics} organizes the metrics in four families, defined formally in App.~\ref{app:metric-definitions}. For each metric, the table specifies whether it uses the features $X$ or the graph structure, and whether it depends on the choice of $\Delta$-Laplacian. 
A metric built on the Laplacian, e.g.\ the RSA through the norm $\|x\|_\Lap = \sqrt{x^\top \Lap x}$, depends on that choice. It takes its own \texttt{laplacian\_kind} argument, which defaults to the one of the coarsening. For instance, a partition that local variation chooses with the self-loop normalized Laplacian can be scored under the combinatorial one.

The table also gives, for each metric, the paper that introduced it and papers that reuse it. The spectral metrics and the energy errors on the features are shared across the literature, while method-specific metrics remain confined to their original paper.

\begin{table}[t]
\centering
\footnotesize
\setlength{\tabcolsep}{4.5pt}
\caption{Metrics of the toolkit, grouped by family. \emph{Source} is the paper that introduced the metric, and \emph{Used by} the other coarsening papers that report it; a dash marks no other user, or an unclear origin. \emph{Feat.}, \emph{Struct.} and \emph{Label} mark the metrics that use the features, the graph structure and the node labels. A check under \emph{Lap.} means that the metric depends on the choice of $\Delta$-Laplacian.}
\label{tab:metrics}
\begin{tabular}{lllcccc}
\toprule
Metric & Source & Used by & Feat. & Struct. & Lap. & Label \\
\midrule
\multicolumn{7}{l}{\emph{Spectral fidelity}} \\
\texttt{rsa}                          & \cite{loukas2019graph}             & \cite{joly2024graph, joly2026core, joly2026taxonomy}     &            & \checkmark & \checkmark & \\
\texttt{mean\_REE}                    & \cite{loukas2019graph}             & \cite{kumar2023featured, kataria2024ugc, cohen2026spectral} &            & \checkmark & \checkmark & \\
\texttt{spectral\_distance}           & \cite{wilson2008study}           & \cite{xia2025gbgc}                                          &            & \checkmark & \checkmark & \\
\midrule
\multicolumn{7}{l}{\emph{Operator reconstruction}} \\
\texttt{reconstruction\_error}        & --                                 & \cite{kumar2023featured, cohen2026spectral}                 &            & \checkmark & \checkmark & \\
\texttt{gromov\_wasserstein\_distortion} & \cite{taveras2026gromov}  & --         &            & \checkmark & \checkmark & \\
\midrule
\multicolumn{7}{l}{\emph{Features}} \\
\texttt{rsa\_features}                & \cite{kumar2023featured}           & \cite{kataria2024ugc}                                       & \checkmark & \checkmark & \checkmark & \\
\texttt{dirichlet\_energy\_error}     & --                                 & \cite{kumar2023featured, cohen2026spectral}                 & \checkmark & \checkmark & \checkmark & \\
\texttt{hyperbolic\_error}            & \cite{bravo2019unifying} & \cite{kumar2023featured, kataria2024ugc, cohen2026spectral} & \checkmark & \checkmark & \checkmark & \\
\texttt{inner\_product\_error}        & \cite{cohen2026spectral}           & --                                                          & \checkmark & \checkmark & \checkmark & \\
\texttt{convolution\_matching\_error} & \cite{dickens2024graph}            & --                                                          & \checkmark & \checkmark &            & \\
\texttt{wasserstein\_feature\_distance}  & \cite{ma2021unsupervised} & --         & \checkmark &            &            & \\
\midrule
\multicolumn{7}{l}{\emph{Inside the supernodes}} \\
\texttt{max\_acc}                     & \cite{joly2024graph}               & --                                                          &            &            &            & \checkmark \\
\texttt{partition\_density\_quality}  & \cite{xia2025gbgc}                 & --                                                          &            & \checkmark &            & \\
\bottomrule
\end{tabular}
\end{table}

\paragraph{Spectral fidelity.} These metrics work on the spectrum of $\Lap$.
\begin{itemize}[leftmargin=*, itemsep=1pt]
\item \texttt{rsa}, the \emph{Restricted Spectral Approximation} of Loukas~\cite{loukas2019graph}: the worst-case error $\|x - \Pi x\|_\Lap$ over the low-frequency signals, spanned by the first eigenvectors of $\Lap$.
\item \texttt{mean\_REE}: the mean relative error between the smallest eigenvalues of $\Lap$ and those of $\Lapc$.
\item \texttt{spectral\_distance}: the gap between the whole spectrum of $\Lapc$ and the smallest eigenvalues of $\Lap$.
\end{itemize}

\paragraph{Operator reconstruction.} These metrics compare an operator of the graph with its approximation through $\Pi$.
\begin{itemize}[leftmargin=*, itemsep=1pt]
\item \texttt{reconstruction\_error}: the Laplacian $\Lap$ against $\Pi^\top \Lap \Pi$.
\item \texttt{gromov\_wasserstein\_distortion}: a structure matrix of the graph, the adjacency or a normalized version of it, against its block means over the pairs of supernodes.
\end{itemize}

\paragraph{Features.} These metrics compare $X$ with the coarsened features.
\begin{itemize}[leftmargin=*, itemsep=1pt]
\item \texttt{rsa\_features}: the RSA error of each feature column.
\item \texttt{dirichlet\_energy\_error}: the error on the energy $\tr(X^\top \Lap X)$, and \texttt{hyperbolic\_error} the same comparison in a scale-invariant form.
\item \texttt{inner\_product\_error}: the error on the full Gram matrix $X^\top \Lap X$, the objective of IPE.
\item \texttt{convolution\_matching\_error}: the error on the output of one graph convolution, the objective of ConvMatch.
\item \texttt{wasserstein\_feature\_distance}: a transport cost between the feature distributions of the nodes and of the supernodes.
\end{itemize}

\paragraph{Inside the supernodes.} These metrics use the partition itself.
\begin{itemize}[leftmargin=*, itemsep=1pt]
\item \texttt{max\_acc}: the fraction of nodes that have the majority label of their supernode, a training-free upper bound on the accuracy of a model that predicts one label per supernode~\cite{joly2024graph}.
\item \texttt{partition\_density\_quality}: the score GBGC uses to decide whether to split a ball.
\end{itemize}

\section{A synthetic check}
\label{sec:sbm}

\begin{figure}[htbp]
\centering
\includegraphics[width=\linewidth]{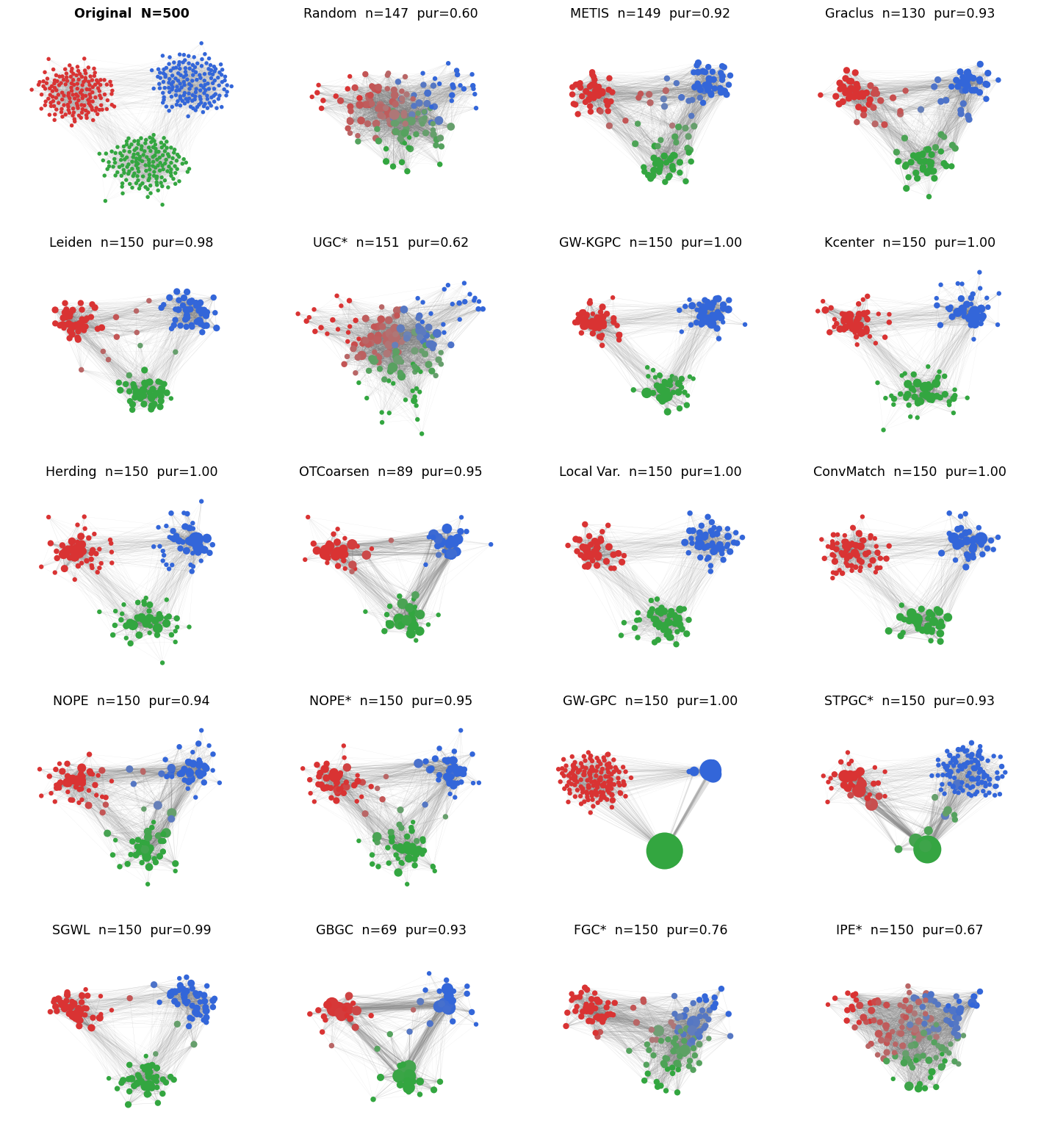}
\caption{Coarsened graph produced by each method on a three-community SBM ($\Nnodes = 500$, $p_{\mathrm{in}} = 0.12$, $p_{\mathrm{out}} = 0.005$), at the target ratio $r = 0.7$. The top-left panel is the original graph. UGC, STPGC, FGC and IPE, marked with *, use hyperparameters tuned for this graph (App.~\ref{app:hp-sensitivity}), and the others use their paper defaults. The panel NOPE* is the variant NOPE$^{*}$ (Sec.~\ref{sec:methods}). The same figure at $r = 0.5$ and $r = 0.9$ is in App.~\ref{app:sbm-ratios}.}
\label{fig:sbm}
\end{figure}

We propose a toy experiment on a stochastic block model (SBM) to check that the implementations behave as expected on a simple graph with a known structure, and to illustrate the differences between methods. 
We draw an SBM of 500 nodes with three communities of roughly equal size, a dense intra-community edge probability, and a sparse inter-community probability. 
We attach ten node features drawn as $X \sim \mathcal{N}(0, \Lap^\dagger)$, the convention Kumar et al.~\cite{kumar2023featured} use for each graph of their experiments that has none of its own. 
At a moderate ratio, a good coarsening puts the nodes of a community in the same supernodes, a weaker requirement than exact community recovery~\cite{abbe2018community}.
We measure the purity \texttt{max\_acc} of Sec.~\ref{sec:metrics}, with the communities as labels.
Besides the purity, Fig.~\ref{fig:sbm} shows whether the three-block structure survives. 
Each supernode is drawn at the mean position of its members in the layout of the original graph, with a size that grows with its number of nodes, so a coarsening that respects the communities shows three well-separated groups of supernodes. 
A supernode that mixes communities falls between them, and is greyed accordingly.

 Most methods preserve the planted communities. METIS, Graclus, Leiden, GW-KGPC, Kcenter, Herding, OTCoarsening, local variation, ConvMatch, NOPE and SGWL merge within communities, with a purity above $0.9$. They mostly differ in how they spread the reduction. A community is coarsened more when its nodes fall in fewer, larger supernodes. Some methods coarsen the three communities evenly, while GW-GPC and STPGC concentrate the reduction on one community first. The hashing and dense-optimization methods struggle. UGC mixes communities through its hash of features and adjacency, and FGC no longer respects them once its soft lifting matrix is discretized by a row-wise argmax. IPE recovers the communities only when its Grassmann rank is set well below the requested number of supernodes, and it then returns far fewer supernodes than requested (App.~\ref{app:sbm-ipe}, Fig.~\ref{fig:ipe-k}).
The features here are smooth random signals, which depend on the communities only through the graph that generated them. UGC and FGC may do better on real graphs, where the features often contain class information.

App.~\ref{app:rgg} repeats this check on a random geometric graph, whose nodes are points drawn in the unit square.

\FloatBarrier
\section*{Acknowledgments}
We acknowledge the funding of France 2030, PEPR IA SHARP, ANR-23-PEIA-0008. We thank Nicolas Keriven for his advice and his comments on the paper, Esteban Pesnel for help in reviewing the code, and Adarsh Jamadandi for a careful reading of the paper.

\bibliographystyle{plainnat}
\bibliography{refs}

\appendix
\raggedbottom
\clearpage

\section*{Appendix contents}
{\small
\setlength{\tabcolsep}{4pt}
\renewcommand{\arraystretch}{1.15}
\begin{tabular}{@{}p{0.6cm}p{11.9cm}r@{}}
A & \hyperref[app:method-catalogue]{Method catalogue} & \pageref{app:method-catalogue}\\
 & \quad \hyperref[app:cat-clustering]{Global clustering} $\cdot$ \hyperref[app:cat-selection]{Representative selection} $\cdot$ \hyperref[app:cat-contraction]{Greedy contraction} $\cdot$ \hyperref[app:cat-splitting]{Recursive splitting} $\cdot$ \hyperref[app:cat-dense]{Dense optimization} & \\
B & \hyperref[app:metric-definitions]{Metric definitions} & \pageref{app:metric-definitions}\\
 & \quad \hyperref[app:met-spectral]{Spectral fidelity} $\cdot$ \hyperref[app:met-operator]{Operator reconstruction} $\cdot$ \hyperref[app:met-features]{Features} $\cdot$ \hyperref[app:met-supernodes]{Inside the supernodes} $\cdot$ \hyperref[app:cross-scoring]{Scoring on Cora and PubMed} & \\
C & \hyperref[app:pseudocodes]{Method remarks and extensions} & \pageref{app:pseudocodes}\\
 & \quad \hyperref[app:laplacian-table]{Operator conventions} $\cdot$ \hyperref[app:extensions]{Summary of our extensions} & \\
 & \quad \hyperref[app:otcoarsening-variants]{OTCoarsening} $\cdot$ \hyperref[app:local-variation]{Local Variation} $\cdot$ \hyperref[app:convmatch-pq]{ConvMatch} $\cdot$ \hyperref[app:nope-variants]{NOPE} $\cdot$ \hyperref[app:stpgc-variants]{STPGC} $\cdot$ \hyperref[app:sgwl-entry-points]{SGWL} $\cdot$ \hyperref[app:gbgc-sqrtn]{GBGC} $\cdot$ \hyperref[app:fgc-code]{FGC} $\cdot$ \hyperref[app:ipe-instability]{IPE} & \\
D & \hyperref[app:datasets]{Datasets and preprocessing} & \pageref{app:datasets}\\
E & \hyperref[app:experimental]{Experiments on synthetic graphs} & \pageref{app:experimental}\\
 & \quad \hyperref[app:sbm-ratios]{SBM at other ratios} $\cdot$ \hyperref[app:hp-sensitivity]{Hyperparameter tuning} $\cdot$ \hyperref[app:sbm-ipe]{IPE and the Grassmann rank} $\cdot$ \hyperref[app:rgg]{A random geometric graph} & \\
F & \hyperref[app:details]{Additional details} & \pageref{app:details}\\
 & \quad \hyperref[app:storage]{Storing and reloading a coarsening} & \\
G & \hyperref[app:related]{Related surveys and tooling} & \pageref{app:related}\\
\end{tabular}}

\section{Method catalogue}
\label{app:method-catalogue}
We describe each method in the family order of Sec.~\ref{sec:methods}: first the method as published, then our implementation. Three short items close each entry. \emph{Complexity} gives the asymptotic cost, \emph{Implementation} the backend and the libraries it relies on, and \emph{Scalability} the tier of Table~\ref{tab:method-taxonomy}, measured in Table~\ref{tab:runtime}.

\subsection{Global clustering}
\label{app:cat-clustering}

\paragraph{Random.}
\label{sec:methods:random}
The random baseline assigns each node of the original graph to one of the $\nclust$ supernodes, independently and uniformly at random. It uses neither the adjacency $\Adj$ nor the features $X$. It fixes an uninformative lower bound, so that a metric reported for a coarsening method can be compared with it. We consider two variants. The \textbf{i.i.d.\ uniform} variant leaves cluster sizes as a multinomial draw around $\Nnodes/\nclust$. The \textbf{equal-size} variant shuffles the node index list and splits it into $\nclust$ contiguous chunks of size $\lceil \Nnodes/\nclust \rceil$ or $\lfloor \Nnodes/\nclust \rfloor$. 

The i.i.d.\ variant is the default. Note that it does \emph{not} reach the requested ratio. A uniform draw leaves some supernodes empty. Thus, the realized supernode count is $\nclust (1 - e^{-\Nnodes / \nclust})$ in expectation. On Cora at $r = 0.5$ this returns $1064$ supernodes instead of $1243$, that is $r = 0.57$. The gap shrinks as $r$ grows. The equal-size variant reaches the target exactly.

\emph{Complexity.} $\mathcal{O}(\Nnodes)$ time and memory. \emph{Implementation.} Native torch, GPU. \emph{Scalability.} Very large.

\paragraph{METIS\@.}
\label{sec:methods:metis}
Karypis and Kumar~\cite{karypis1997metis} introduced METIS as a multilevel graph partitioning package. Given a target partition count $\nclust$, METIS minimizes the number of edges cut between the $\nclust$ clusters under a balance constraint,
\begin{equation}
  \min_{V_1, \dots, V_\nclust} \ \tfrac{1}{2} \sum_{k=1}^{\nclust} \sum_{\substack{i \in V_k \\ j \notin V_k}} \Adj_{ij}
  \qquad \text{s.t.} \qquad |V_k| \leq (1 + \tau)\, \tfrac{\Nnodes}{\nclust},
  \label{eq:metis-cut}
\end{equation}
with $\tau$ a small imbalance tolerance, so cluster sizes are close to $\Nnodes / \nclust$. The algorithm proceeds in three phases. First, a \textbf{coarsening} phase contracts edges via heavy-edge matching to produce a small graph. Second, an \textbf{initial partitioning} phase splits this small graph by recursive bisection. Third, an \textbf{uncoarsening} phase lifts the partition back to the original graph and refines it by local swaps at each level.

We wrap the reference \texttt{pymetis} binding with a single hyperparameter, the random seed. METIS enforces an implicit minimum cluster size of roughly three nodes. As a result it returns fewer non-empty clusters than requested at low coarsening ratios. For instance, on the Cora largest connected component ($\Nnodes = 2485$), requesting $r = 0.3$ (i.e.\ $1740$ clusters) returns only $701$ non-empty clusters, a $60\%$ shortfall. The gap closes to $7\%$ at $r = 0.7$ and vanishes at $r \geq 0.9$.

\emph{Complexity.} $\mathcal{O}(E)$ time. \emph{Implementation.} \texttt{pymetis} wraps the reference C library, CPU only. \emph{Scalability.} Very large.

\paragraph{Graclus.}
\label{sec:methods:graclus}
Dhillon, Guan and Kulis~\cite{dhillon2007weighted} introduced Graclus as a multilevel algorithm that minimizes normalized-cut or ratio-cut objectives. For the normalized cut, the objective is
\begin{equation}
  \min_{V_1, \dots, V_\nclust} \ \sum_{k=1}^{\nclust} \frac{\operatorname{cut}(V_k, \bar{V}_k)}{\operatorname{vol}(V_k)},
  \qquad \operatorname{cut}(V_k, \bar{V}_k) = \sum_{\substack{i \in V_k \\ j \notin V_k}} \Adj_{ij},
  \label{eq:ncut}
\end{equation}
with $\operatorname{vol}(V_k) = \sum_{i \in V_k} D_{ii}$ the degree mass of cluster $k$.
 Graclus differs from METIS by targeting a normalized cut objective, which biases the partition toward clusters of comparable degree mass. METIS cuts edges and balances node count.
The original algorithm alternates a heavy-edge matching phase and a refinement phase guided by a weighted kernel $k$-means objective, a relaxation of the normalized cut.

We use the PyG implementation (\texttt{torch\_geometric.nn.pool.graclus}, backed by \texttt{torch-cluster}), which performs the matching phase only, without the refinement. One step matches each node with at most one neighbor, so it reduces the graph by at most half, and the number of supernodes it returns is not chosen. We therefore apply it recursively until the number of supernodes falls below $(1-r)\Nnodes$, which may reduce more than asked.

\emph{Complexity.} $\mathcal{O}(E \cdot L)$ with $L$ the number of levels. \emph{Implementation.} PyG \texttt{graclus}, backed by \texttt{torch-cluster}, GPU. \emph{Scalability.} Very large.

\paragraph{Leiden.}
\label{sec:methods:leiden}
Traag, Waltman and van Eck~\cite{traag2019louvain} introduced Leiden as a refinement of the Louvain community detection algorithm~\cite{blondel2008fast}. Both maximize the \textbf{modularity} of the partition,
\begin{equation}
  \mathrm{Mod}_\gamma = \frac{1}{2E} \sum_{i, j} \Big( \Adj_{ij} - \gamma \frac{d_i d_j}{2E} \Big)\, (\Lift_b \Lift_b^\top)_{ij},
  \label{eq:modularity}
\end{equation}
which compares the edges inside each community with the number expected at random for the same degrees $d_i = D_{ii}$. Here $2E = \mathbf{1}^\top \Adj \mathbf{1}$ is the total edge weight. The entry $(\Lift_b \Lift_b^\top)_{ij}$ is $1$ when nodes $i$ and $j$ are in the same supernode, $0$ otherwise. The resolution $\gamma$ controls the number of communities. Louvain greedily moves each node to the neighboring community that increases $\mathrm{Mod}_\gamma$ the most, then contracts the communities into supernodes and repeats. Some of its communities can be internally disconnected, and Leiden adds a refinement step that splits them into connected parts before contraction.

Leiden takes a resolution and no number of supernodes. We therefore search $\gamma$ by bisection on a log scale over $[10^{-3}, 10^{4}]$, until the number of communities is within one of $(1-r)\Nnodes$. When no resolution reaches the target, the search returns the closest partition it met: on Cora, a request for $1740$ communities ($r = 0.3$) returns $1696$. A fixed \texttt{resolution} can also be passed, which skips the search.

\emph{Complexity.} $\mathcal{O}(E)$ per sweep of local moves, repeated for each resolution. \emph{Implementation.} \texttt{igraph} + \texttt{leidenalg}, CPU only. \emph{Scalability.} Very large.

\paragraph{UGC\@.}
\label{sec:methods:ugc}
Kataria, Kumar and Jayadeva~\cite{kataria2024ugc} introduced Universal Graph Coarsening (UGC) as a hashing-based coarsening method designed to work on both homophilic and heterophilic graphs. Each node $i$ receives an augmented feature vector that concatenates its feature row and its adjacency row, mixed by a \textbf{heterophily factor} $\alpha \in [0, 1]$,
\begin{equation}
  F_i = \big[(1-\alpha)\, X_i,\ \alpha\, \Adj_i\big],
  \qquad
  \alpha = \frac{|\{(u, v) \in E : y_u \neq y_v\}|}{|E|},
  \label{eq:ugc}
\end{equation}
i.e.\ the fraction of edges joining nodes of different classes, computed from the label vector $y$. As in the paper, our default computes it on the labels of all nodes, test nodes included. A heterophilic graph then gives more weight to the adjacency, and a homophilic one to the features.

UGC then groups the nodes whose augmented vectors are close, by locality-sensitive hashing. It draws $m$ random directions and projects every $F_i$ on each of them. It then cuts each projection into bins of width $w$, so that two close vectors tend to fall in the same bin. Each node goes to the bin it falls in most often over the $m$ projections, and each bin becomes a supernode. The bin width sets the number of supernodes; our implementation searches it by bisection until this number meets the target, within $5\%$. We keep the defaults of the official repository, $m = 500$ directions drawn uniformly with dot-product hashing.

\emph{Complexity.} $\mathcal{O}(\Nnodes m (d + \bar{d}))$, with $\bar{d}$ the average degree. \emph{Implementation.} Sparse projections of $X$ and $\Adj$ (App.~\ref{app:extensions}), GPU. \emph{Scalability.} Very large. \emph{Label-aware.} Yes, via the heterophily factor $\alpha$. 

\paragraph{GW KGPC\@.}
\label{sec:methods:gw}
Taveras et al.~\cite{taveras2026gromov} look at coarsening through the Gromov-Wasserstein (GW) distance. A graph is a measure network $(S, \mu)$, with $S$ a matrix representing the graph, e.g.\ its adjacency, and $\mu$ a measure on the nodes. For a binary lifting $\Lift$, they build $C = \operatorname{diag}(\mu)\Lift\operatorname{diag}(1/\Lift^\top\mu)$ and score the coarsening by the distortion
\begin{equation}
    D = \big\|(S - \Pi S \Pi)\odot(\mu\mu^\top)^{1/2}\big\|_F^2, \qquad \Pi = \Lift C^\top.
    \label{eq:gw}
\end{equation}
$D$ is the Gromov-Wasserstein cost between the graph and its coarsened graph $C^\top S C$, when the transport plan sends each node to its supernode.
The partition plays the role of the transport plan, restricted to plans that do not split a node into multiple supernodes.
For a uniform $\mu$, the paper default, $C$ is the Moore-Penrose reduction $\Proj = \Lift^{+}$, $\Pi$ averages a signal over each supernode, and $D$ is proportional to $\|S - \Pi S \Pi\|_F^2$. Note that $\Pi S \Pi = \Lift (\Proj S \Lift) \Proj$, with $\Proj S \Lift$ the coarse propagation of~\cite{joly2024graph}. $D$ thus measures how much the operator $S$ changes through the coarsening, for all signals at once, similarly to ConvMatch for one convolution.

Two algorithms minimize $D$. GPC, a greedy contraction, is described in App.~\ref{sec:methods:gw-gpc}. \textbf{KGPC} computes the distortion of merging each pair of nodes once, on the original graph, and gathers them in a matrix $H \in \R^{\Nnodes \times \Nnodes}$. It then runs a $k$-means on the rows of $H$, and each cluster becomes a supernode. Our implementation provides five choices of $S$ and a degree-normalized $\Pi$ in addition to the uniform one (App.~\ref{app:extensions}).

\emph{Complexity.} KGPC builds $H$ once in $\mathcal{O}(\Nnodes^3)$, then runs $T$ iterations of $k$-means in $\mathcal{O}(T \Nnodes^2 \nclust)$. It stores $S$ densely, in $\mathcal{O}(\Nnodes^2)$ memory. \emph{Implementation.} Native torch, GPU, with a torch $k$-means. \emph{Scalability.} Medium.

\subsection{Representative selection}
\label{app:cat-selection}

\paragraph{Kcenter.}
\label{sec:methods:kcenter}
Gonzalez~\cite{gonzalez1985clustering} proposed a greedy algorithm for the \textbf{$k$-center} problem, which selects $\nclust$ centers so that every point lies as close as possible to its nearest center. It starts from one point and repeatedly adds the point farthest from the centers already selected (farthest-point sampling), which is within a factor two of the optimum. We run it on the nodes in an embedding space, then assign every other node to its nearest center, so that each center and the nodes around it form a supernode. 

The Kcenter and Herding baselines of the condensation literature~\cite{jin2022graph} are \emph{supervised}: they train a GCN on the original graph and select nodes class by class. Ours use neither labels nor training, so the two are not comparable.

The embedding space is $H = \Prop^K X$, where $\Prop$ is the self-loop normalized propagation of Sec.~\ref{sec:notation} and $K$ is the number of Simplified-Graph-Convolution~\cite{wu2019simplifying} propagation steps. Larger $K$ smooths the node representations more before the selection runs: $K = 0$ falls back to the raw features $X$. The distance is the Euclidean distance in $H$. The graph enters only through $\Prop^K$, so the supernodes follow the topology up to $K$ hops. We default to $K = 2$ propagation steps, matching the SGC-2 embedding of ConvMatch.

\emph{Complexity.} $\mathcal{O}(K E d)$ for the $K$ propagation steps and $\mathcal{O}(\Nnodes \nclust d)$ for the $\nclust$ steps of farthest-point sampling.
\emph{Implementation.} Native torch, GPU.
\emph{Scalability.} Very large at high ratios, where $\nclust$ is small; at a fixed ratio the selection is quadratic in $\Nnodes$.

\paragraph{Herding.}
\label{sec:methods:herding}

Herding, introduced by Welling~\cite{welling2009herding}, selects a small subset of points whose mean is as close as possible to the mean $\mu$ of the whole set. It is greedy, and each step adds the point that brings the mean of the selected points closest to $\mu$,
\begin{equation}
  h_t = \arg\min_{h} \Big\| \mu - \tfrac{1}{t} \Big( \sum_{s < t} h_s + h \Big) \Big\|_2 .
  \label{eq:herding}
\end{equation}
Herding thus selects points that are representative on average, which favors the dense regions.

We use the same embedding $H = \Prop^K X$ as Kcenter ($K = 2$ by default), select $\nclust$ pivots by herding on $H$, and assign every other node to its nearest pivot. Two nodes with the same closed neighborhood have the same row of $H$. When both are selected, every node closest to one is as close to the other, so one pivot keeps no node and Herding returns fewer supernodes than requested (App.~\ref{app:rgg}).

\emph{Complexity.} $\mathcal{O}(K E d)$ for the propagation and $\mathcal{O}(\Nnodes \nclust d)$ for the $\nclust$ steps of herding. \emph{Implementation.} Native torch, GPU. \emph{Scalability.} Very large at high ratios.

\paragraph{OTCoarsening.}
\label{sec:methods:otcoarsening}
Ma and Chen~\cite{ma2021unsupervised} learn a coarsening without labels, by training a graph neural network to minimize a transport loss. A GCN layer scores each node, $\alpha = \sigma(\hat{\Adj} X W_\alpha)$, and the $\nclust$ nodes of highest score become the supernodes. The lifting $\Lift \in \R^{\Nnodes \times \nclust}$ keeps the columns of $\hat{\Adj}$ at these nodes, weighted by their scores and normalized per row, so each node is spread over the selected nodes within one hop. The coarsened graph is $\Adjc = \Lift^\top \Adj \Lift$. Its features $X_c$ come from an encoder-decoder: a GCN encodes $X$ on the original graph, $\Lift^\top$ pools the encoding onto the supernodes, and a second GCN decodes it on $\Adjc$. The loss is the entropic Wasserstein distance between the rows of $X$ and those of $X_c$,
\begin{equation}
    W_\gamma(X, X_c) = \min_{T} \sum_{i,j} T_{ij} \|X_i - X_{c,j}\|_p^p - \gamma E(T),
    \label{eq:otcoarsening}
\end{equation}
over the plans $T$ whose rows and columns sum to the node masses, with $E$ the entropy. Unlike in the Gromov-Wasserstein methods, the transport cost compares feature vectors. The graph structure enters it only through the decoder that produces $X_c$. A fixed number of Sinkhorn iterations computes $T$, and the loss is differentiated through them. The score layer and the two GCNs are thus trained together, so that the features of the supernodes cover the distribution of the original ones.

After training, our implementation sends each node to the selected node of largest weight $\hat{\Adj}_{ij}\alpha_j$ in its row of $\Lift$, i.e.\ to the best-scored selected node among itself and its neighbors. A node with no selected node within one hop has an empty row. By default it becomes a supernode of its own, and the coarsened graph is coarsened again until $\nclust$ supernodes remain (App.~\ref{app:otcoarsening-variants}). The paper needs no such step, since it targets graph classification, where $\Lift$ only pools the graph and such a node drops out. The same appendix lists the points where the released code differs from the paper.

\emph{Complexity.} $\mathcal{O}(\Nnodes^2 d)$ per epoch, since the paper builds the graph convolutions and $\Lift$ on the dense adjacency, plus $\mathcal{O}(k \Nnodes \nclust)$ for $k$ Sinkhorn iterations. A sparse implementation would bring the first term down to $\mathcal{O}(E d)$; we keep the published one. \emph{Implementation.} Native torch, GPU. \emph{Scalability.} Small; the dense $\Nnodes \times \Nnodes$ matrices run out of memory on PubMed.

\subsection{Greedy contraction}
\label{app:cat-contraction}

\paragraph{Local Variation.}
\label{sec:methods:local-variation}
Loukas~\cite{loukas2019graph} proposes the local variation methods, which aim to find the partition which best preserves the low-frequency signals of the graph. Loukas measures it by the \textbf{Restricted Spectral Approximation} (RSA) constant over the span $R$ of the first $K$ eigenvectors of $\Lap$ (we default to $K = 100$),
\begin{equation}
  \epsilon_{R} = \sup_{x \in R,\ \|x\|_\Lap = 1} \|x - \Lift \Proj x\|_\Lap,
  \qquad \|x\|_\Lap = \sqrt{x^\top \Lap x}.
  \label{eq:loukas-rsa}
\end{equation}
The constant depends on both $\Lift$ and $\Proj$. Loukas takes $\Proj = \Lift^{+}$ at each level, so once the Laplacian is fixed it depends on the partition only.

Minimizing Eq.~\eqref{eq:loukas-rsa} directly over all partitions is out of reach. Loukas instead builds the coarsening in levels and bounds the error of each one. The RSA of the whole descent is at most $\prod_{\ell} (1 + \sigma_\ell) - 1$, with $\sigma_\ell$ the error of level $\ell$. Each $\sigma_\ell^2$ is in turn bounded by a sum of local terms, one per contraction set $\mathcal{C}$.  A candidate set can therefore be scored on its own, by its \emph{local variation cost}
\begin{equation}
  \mathrm{cost}_\ell(\mathcal{C}) = \frac{\|\Pi_{\mathcal{C}}^{\perp} B_{\ell-1}\|^2_{\Lap_{\mathcal{C}}}}{|\mathcal{C}| - 1},
  \label{eq:loukas-cost}
\end{equation}
where $B_{\ell-1}$ is the preserved basis reduced to level $\ell - 1$, $\Pi_{\mathcal{C}}^{\perp}$ removes the mean of a signal over $\mathcal{C}$, weighted by the masses of the supernodes in our default (App.~\ref{app:local-variation}), and $\Lap_{\mathcal{C}}$ is the Laplacian of the edges touching $\mathcal{C}$. The denominator counts the nodes the contraction removes.

Even with this decoupling, choosing the cheapest family of disjoint candidate sets is NP-hard, by reduction to maximum-weight independent set. Loukas therefore proposes a greedy algorithm. At each level, it builds a family of candidate sets, either each edge (\textbf{variation edges}, our default) or each node with its neighbors (\textbf{variation neighborhood}). 
It then scans the candidates by increasing cost, and contracts each one that shares no node with a set already accepted. A neighborhood that meets such a node loses it, is costed again and goes back in the queue, as in the algorithm of~\cite{loukas2019graph}. The scan stops once a fraction \texttt{n\_e\_percent} of the current nodes is contracted. The process repeats on the coarsened graph. Loukas reduces the basis from one level to the next with the pseudoinverse of each level. Our default chooses the reduction of each level so that their product over the levels is $\Lift^+$ of the final partition, for each $\Delta$ (App.~\ref{app:local-variation}). The reduction of Loukas is available as an option. The toolkit implements the modified cost for each $\Delta$-Laplacian. Our implementation also adds an absorption pass for graphs where the disjoint matching stalls (App.~\ref{app:local-variation}).

\emph{Complexity.} $\mathcal{O}(K E + \Nnodes K^2)$ for the initial eigensolve of the $K$ preserved eigenvectors, then $\mathcal{O}(K E + K^3)$ per level over $\mathcal{O}(\log \tfrac{1}{1-r})$ levels. On the combinatorial Laplacian the eigensolve switches to shift-invert, whose factorization can cost far more on large graphs. \emph{Implementation.} \texttt{scipy.sparse} + ARPACK, edge costs vectorized in closed form (App.~\ref{app:local-variation}), \texttt{numba} for the neighborhood cost, CPU only.
\emph{Scalability.} Very large.

\paragraph{ConvMatch.}
\label{sec:methods:convmatch}
Dickens et al.~\cite{dickens2024graph} introduced a method to look for the partition that best preserves the output of one graph propagation. With $\hat{S} = \hat{D}^{-1/2} \hat{A} \hat{D}^{-1/2}$ the propagation of a GCN layer and $\Sc^{\mathrm{H}}$ the same operator rebuilt on the coarsened graph, with the size and the internal edges of each supernode as self-loops, the objective is
\begin{equation}
  \min_{\Lift} \big\| \Lift \Sc^{\mathrm{H}} \Proj X - \hat{S} X \big\|_{1,1},
  \label{eq:convmatch}
\end{equation}
over binary liftings $\Lift$, with $\Proj = \Lift^{+}$. This is the \texttt{dickens} variant of the convolution-matching error of App.~\ref{app:metric-definitions}.

As for local variation, minimizing Eq.~\eqref{eq:convmatch} over all partitions is intractable, and ConvMatch merges pairs of supernodes greedily, level by level. At each level, it embeds the current graph with an SGC~\cite{wu2019simplifying} of depth $K$, draws candidate pairs among the $k$ nearest neighbors of each node in that embedding, and merges the cheapest non-overlapping pairs. The cost of a pair $(u, v)$ is the change of the coarse outputs that merging them causes at the current level. Computing it exactly takes $\mathcal{O}((d_u + d_v)\, d)$, since the merge changes the output of every neighbor of $u$ and $v$. \textbf{A-ConvMatch} replaces it by an upper bound that only uses quantities local to $u$ and $v$, computed in $\mathcal{O}(d)$ (Theorem~1 of the paper). Our implementation follows the paper by default. It adds two other pairs $(\Proj, \Lift)$ inside the cost and a larger candidate set, both detailed in App.~\ref{app:convmatch-pq}.

\emph{Complexity.} $\mathcal{O}(K E d)$ for the SGC embedding at each level, plus $\mathcal{O}(d)$ per candidate pair for A-ConvMatch and $\mathcal{O}((d_u + d_v)\, d)$ for ConvMatch. \emph{Implementation.} Torch, GPU for the embedding and the pair scoring; the candidate search runs \texttt{scipy} \texttt{cKDTree} and \texttt{sklearn} PCA on CPU. \emph{Scalability.} Very large.

\paragraph{NOPE\@.}
\label{sec:methods:nope}
Bai et al.~\cite{bai2026rethinking2} merge the pair whose contraction disturbs their shared neighborhood the least, a principle they call \textbf{non-selfishness}. They measure it by the \textbf{neighborhood interference index} of a candidate pair $(p, q)$,
\begin{equation}
    \mathcal{I}_{pq} = \sum_{i \in (\Ncal_p \cup \Ncal_q) \setminus \{p, q\}} \frac{|v_p^c| |v_q^c|}{|v_p^c| + |v_q^c|} \, (s_{ip} - s_{iq})^2,
    \label{eq:nope}
\end{equation}
with $s_{ij} = \langle x_i, x_j \rangle$ the dot product of the node features, $\Ncal_p$ the neighbors of $p$, and $|v_p^c|$ the size of the current supernode of $p$. A candidate pair is always adjacent, so $p$ belongs to $\Ncal_q$ and $q$ to $\Ncal_p$. The released code removes both from the sum, and we follow it. As for local variation and ConvMatch, the algorithm is greedy: it pops the pair of smallest $\mathcal{I}$ from a priority queue, merges it, and re-scores the pairs around the new supernode.

Computing Eq.~\eqref{eq:nope} exactly costs $\mathcal{O}(\bar{d} d)$ per pair, since the sum runs over the union of the two neighborhoods. \textbf{NOPE$^{*}$} replaces it by its expectation under a local isotropy assumption: the features of the neighbors of a node are i.i.d.\ with covariance $\sigma^2 I_d$. The sum then has a closed form,
\begin{equation}
    \mathbb{E}[\mathcal{I}_{pq}] = \frac{|v_p^c| |v_q^c|}{|v_p^c| + |v_q^c|} \, |\Ncal_r| \, \sigma^2 \, \| x_p - x_q \|^2, \qquad |\Ncal_r| = |\Ncal_p \cup \Ncal_q| - 2,
    \label{eq:nope-star}
\end{equation}
computed in $\mathcal{O}(d)$ per pair. Bai et al.\ recommend NOPE at low to moderate ratios, and NOPE$^{*}$ when scalability matters more. Our implementation provides three variants: \texttt{nope} (Eq.~\eqref{eq:nope}), \texttt{nope\_star}, and \texttt{nope\_repo}, the merge routine of the official repository, which differs from Eq.~\eqref{eq:nope} (App.~\ref{app:nope-variants}). It also generalizes the similarity to $s_{ij}(M) = \langle (MX)_i, (MX)_j \rangle$ for five choices of the matrix $M$, the paper taking $M = I$ (App.~\ref{app:laplacian-table}).

\emph{Complexity.} $\mathcal{O}(\Nnodes \bar{d}^2 d)$ for NOPE and $\mathcal{O}(\Nnodes \bar{d} d)$ for NOPE$^{*}$, with $\bar{d}$ the average degree, which grows as supernodes merge. \emph{Implementation.} Adjacency as \texttt{list[set[int]]} and a priority queue with lazy deletion, CPU only. \emph{Scalability.} Very large through NOPE$^{*}$.

\paragraph{GW GPC\@.}
\label{sec:methods:gw-gpc}
GPC is the greedy algorithm of Taveras et al.~\cite{taveras2026gromov} for the distortion $D$ of Eq.~\eqref{eq:gw} (App.~\ref{sec:methods:gw}). At each step, it computes the distortion of merging each pair of nodes of the current graph, adjacent or not, merges the pair of least distortion, and updates $S$ to $C^\top S C$. It repeats until $\nclust$ supernodes remain. Our implementation provides the same choices of $S$ and $\Pi$ as for KGPC.

\emph{Complexity.} GPC performs $\Nnodes - \nclust$ merges, each scoring all $\mathcal{O}(\Nnodes^2)$ pairs at $\mathcal{O}(\Nnodes)$ each, thus $\mathcal{O}(\Nnodes^4)$ in total. It stores $S$ densely, in $\mathcal{O}(\Nnodes^2)$ memory. \emph{Implementation.} Native torch, GPU. \emph{Scalability.} Small.

\paragraph{STPGC\@.}
\label{sec:methods:stpgc}
Wu et al.~\cite{wuscalable} remove nodes and edges while preserving the topology of the graph, through three local operations on the closed neighborhood $N[u] = \{u\} \cup N(u)$.

\textbf{Strong collapse.} A node $u$ is dominated by a neighbor $v$ if $N[u] \subseteq N[v]$, and it is then merged into $v$. This preserves the topology of the graph, i.e.\ its connected components and its cycles.

\textbf{Edge collapse.} An edge $(x, y)$ is dominated by a common neighbor $v \notin \{x, y\}$ if $N[x] \cap N[y] \subseteq N[v]$, and it is then removed.

\textbf{Neighborhood coning.} When neither applies, the algorithm adds a few edges around a low-degree node $u$ so that one of its neighbors dominates it, then collapses $u$.

All three operations skip the high-degree nodes: they only consider a node of degree at most $\theta_1$, or an edge whose endpoints have degrees summing to at most $2\theta_1$. This threshold bounds the cost of the dominance checks.

The three operations run in a loop of two phases. The \textbf{exact phase} applies them as stated, and preserves the topology of the graph. The \textbf{approximate phase} takes over once the exact one stalls. It no longer adds edges, and it relaxes the strong collapse: $u$ may be merged into $v$ when at most $\rho$ of its neighbors lie outside $N[v]$, and $\rho$ grows by one whenever a round removes fewer than $\theta_2$ nodes, $1\%$ of $\Nnodes$ by default. $\theta_1$ grows with it, as in the released code (App.~\ref{app:stpgc-variants}). This phase reaches any ratio, but it no longer preserves the topology exactly.

The paper uses the train and validation labels in all three operations: it tries the dominators of the same label first, and it keeps the dominated edges between two nodes of the same label. The released code uses the labels of all nodes. Our implementation follows the paper by default, and \texttt{label\_guide=False} gives a label-free run. It leaves out the DropEdge that the paper applies to the coarsened graph before training (App.~\ref{app:stpgc-variants}).

\emph{Complexity.} $\mathcal{O}(E \theta_1^2 + \Nnodes \theta_1^2 \bar{d}^2)$, with $\bar{d}$ the average degree. \emph{Implementation.} CPU; adjacency as \texttt{list[set[int]]}. \emph{Scalability.} Very large. \emph{Label-aware.} Yes, for \texttt{stpgc} (train and validation labels).

\subsection{Recursive splitting}
\label{app:cat-splitting}

\paragraph{SGWL\@.}
\label{sec:methods:sgwl}
Xu et al.~\cite{xu2019scalable} partition a graph by matching it, in the Gromov-Wasserstein sense, with a target graph of $K$ isolated nodes with self-loops, one per part. A transport plan $T \in \R^{\Nnodes \times K}$ gives the mass $T_{ij}$ that node $i$ sends to part $j$: each row sums to the mass $\mu_{s,i}$ of node $i$, and each column to the mass $\mu_{t,j}$ of part $j$. Where GW GPC only compares hard plans, SGWL optimizes a soft $T$ by proximal gradient steps (Eq.~3 of the paper),
\begin{equation}
\begin{split}
    T^{(n+1)} &= \arg\min_{T} \sum_{i,j} M^{(n)}_{ij} T_{ij} + \beta\, \mathrm{KL}\big(T \,\|\, T^{(n)}\big), \\
    M^{(n)}_{ij} &= \sum_{k,l} \big(\Adj_{ik} - \Adj_{t,jl}\big)^2\, T^{(n)}_{kl},
\end{split}
\label{eq:sgwl}
\end{equation}
with $\Adj_t = \operatorname{diag}(\mu_t)$ the adjacency of the target. $M^{(n)}$ is the Gromov-Wasserstein cost with one copy of $T$ frozen at the current plan, which makes it linear in $T$. The KL term keeps each new plan close to the previous one, with $\beta$ as the step size. It also makes each step solvable by Sinkhorn iterations. Expanding the square gives
\begin{equation}
    M^{(n)}_{ij} = \sum_k \mu_{s,k} \Adj_{ik}^2 + \mu_{t,j}^3 - 2\, \mu_{t,j} \sum_k \Adj_{ik} T^{(n)}_{kj}.
\end{equation}
The first term only depends on $i$, the second only on $j$. These two terms do not change the minimizer, since the rows and columns of $T$ have fixed sums. Only the last term matters: $\sum_k \Adj_{ik} T^{(n)}_{kj}$ is the mass that the neighbors of $i$ send to part $j$, so each step pulls a node towards the part of its neighbors. The partition is taken from the final plan, each node going to the part that receives most of its mass, $\Lift = \operatorname{hardmax}(T)$. To scale, the method splits recursively with $K = 2$, so the partition is built by successive binary cuts.

Our implementation stops the recursion once $\nclust$ parts exist, by always splitting the largest part, where the paper fixes a depth and the code a part size (App.~\ref{app:sgwl-entry-points}).  We also provide the GCN propagation operator as a variant of $\Adj$ (App.~\ref{app:extensions}).

\emph{Complexity.} $\mathcal{O}(\Nnodes(E + \Nnodes))$. With $K = 2$ and a fixed number of Sinkhorn iterations, a proximal step on a part of $\Nnodes_{\text{sub}}$ nodes and $E_{\text{sub}}$ edges costs $\mathcal{O}(E_{\text{sub}} + \Nnodes_{\text{sub}})$. Each split runs $\Nnodes_{\text{sub}}$ steps, as in the authors' code, so the first split dominates.
\emph{Implementation.} GPU. \emph{Scalability.} Large; the $\Nnodes E$ time excludes it from the very large tier.

\paragraph{GBGC\@.}
\label{sec:methods:gbgc}
Contrary to most of the other methods we present here, Xia et al.~\cite{xia2025gbgc} coarsen by constructing the coarsened graph from a few supernodes and refining it instead of contracting the original graph. They first cover the graph with about $\sqrt{\Nnodes}$ \textbf{granular-balls}, grown by breadth-first search from the nodes of highest degree. 
Each ball is then split in two by the same breadth-first search, started from its two nodes of highest degree. 
The split is kept if $q(B_1) + q(B_2) > q(B)$, where $q$ measures how tightly the nodes of a ball $B$ are connected,
\begin{equation}
    q(B) = \frac{E_B}{\Nnodes_B} + \frac{\sum_{i \neq j \neq k \in B} \Adj_{ij} \Adj_{jk} \Adj_{ki}}{\sum_{i \neq j \neq k \in B} \Adj_{ij} \Adj_{ik}},
    \label{eq:gbgc}
\end{equation}
where $E_B$ and $\Nnodes_B$ count the edges and the nodes inside $B$.
The first term is the number of internal edges per node. The second is the transitivity: among the pairs of neighbors of a node, the fraction that are themselves connected. The splits recurse until none is accepted, and each final ball becomes a supernode.

The method has no ratio parameter. Our implementation reaches a requested $\nclust$ by stopping the splits once the ball count meets it, so the initial balls set a floor on $\nclust$. On graphs with hubs, the breadth-first initialization returns many more balls than $\sqrt{\Nnodes}$, which caps the reachable ratio. A later work of the same group~\cite{wang2026efficient} lets METIS set the ball count instead, which returns it exactly. We therefore report two versions: \texttt{gbgc}, with the breadth-first initialization of Xia et al., and \texttt{gbgc\_metis}, with the METIS one and the rest of GBGC unchanged. The other two changes of the later work, a mean acceptance rule and a quality without the transitivity, are available as options, as is a factor that softens the acceptance rule (App.~\ref{app:gbgc-sqrtn}).

\emph{Complexity.} $\mathcal{O}(\sqrt{\Nnodes}\,(\Nnodes + E))$. A ball has at most $\sqrt{\Nnodes}$ nodes, so each node and each edge goes through at most $\sqrt{\Nnodes}$ splits.
\emph{Implementation.} CPU; adjacency as \texttt{list[set[int]]}, \texttt{pymetis} for the METIS initialization. \emph{Scalability.} Medium for \texttt{gbgc}, with a reachable ratio that depends on the graph; very large for \texttt{gbgc\_metis}.

\subsection{Dense optimization}
\label{app:cat-dense}

\paragraph{FGC\@.}
\label{sec:methods:fgc}
Kumar et al.~\cite{kumar2023featured} optimize the structure and the features of the coarsened graph jointly.
They relax the lifting matrix to a soft $\Lift \in \R^{\Nnodes \times \nclust}$, $\Lift \geq 0$, and minimize
\begin{equation}
    f(\Lift, X_c) = \tr(X_c^\top \Lift^\top \Lap \Lift X_c) - \gamma \log\det(\Lift^\top \Lap \Lift + J) + \tfrac{\alpha}{2} \| X - \Lift X_c \|_F^2 + \tfrac{\lambda}{2} \|\Lift^\top\|_{1,2}^2,
    \label{eq:fgc}
\end{equation}
with $X_c \in \R^{\nclust \times d}$ the coarsened features, $J = \tfrac{1}{\nclust} \mathbf{1} \mathbf{1}^\top$, and $\gamma, \alpha, \lambda$ the weights of the last three terms, the hyperparameters of the method. The first term is the Dirichlet energy of $X_c$ on the coarsened Laplacian $\Lapc = \Lift^\top \Lap \Lift$, so it asks the coarsened features to be smooth. The log-determinant keeps $\Lapc + J$ well conditioned, i.e.\ the coarsened graph connected. The third term ties $\Lift X_c$ to the original features. The last one, $\|\Lift^\top\|_{1,2}^2 = \sum_i (\sum_j \Lift_{ij})^2$, is a \textbf{group-sparsity} penalty on the rows of $\Lift$. With $\Lift \geq 0$, it pushes each node towards a single supernode, a soft form of the partition constraint (Sec.~3.5 of the paper).

The paper alternates a closed-form update of $X_c$ with a projected gradient step on $\Lift$. The partition is the row-wise argmax of the final $\Lift$. 
Note that the soft $\Lift$ is dense, with $\Nnodes \nclust = (1 - r) \Nnodes^2$ entries. At a fixed ratio, its memory grows quadratically with the size of the graph. Thus, FGC scales only at ratios high enough to keep $\nclust$ small.

Our implementation provides the two solvers of the released code through \texttt{variant} (App.~\ref{app:fgc-code}).
FGC is one of the four methods whose hyperparameters we tune on the block model (App.~\ref{app:hp-sensitivity}).

\emph{Complexity.} $\mathcal{O}(\Nnodes \nclust^2)$ time per iteration and $\mathcal{O}(\Nnodes \nclust)$ memory for the dense $\Lift$. \emph{Implementation.} Native torch, GPU. \emph{Scalability.} Medium, large at high ratios.

\paragraph{IPE\@.}
\label{sec:methods:ipe}
Cohen \& Talmon~\cite{cohen2026spectral} start from FGC and change the objective function slightly. While FGC wants each feature column to be reconstructed, IPE asks the inner products between feature columns, $\langle x, y \rangle_\Lap = x^\top \Lap y$, to be preserved, i.e.\ the Gram matrix $X^\top \Lap X$. They keep the soft lifting, the log-determinant and the group sparsity of FGC, and minimize
\begin{equation}
\begin{split}
    f(\Lift) = {} & \| X^\top \Lap X - X_c^\top \Lift^\top \Lap \Lift X_c \|_F^2 - \beta \tr\big(U_k U_k^\top \Lift \Lift^\top\big) \\
    & - \alpha \log\det(\Lift^\top \Lap \Lift + J) + \lambda \|\Lift^\top\|_{1,2}^2, \qquad X_c = \Lift^{+} X,
\end{split}
\label{eq:ipe}
\end{equation}
with $\alpha, \beta, \lambda$ the hyperparameters, named as in the paper. Note that $\alpha$ weighs the log-determinant here, where FGC calls it $\gamma$. The first term is the \textbf{inner-product error}. The second, a \textbf{Grassmann similarity}, pulls the column space of $\Lift$ towards the span of $U_k$, the $k$ eigenvectors of $\Lap$ of smallest eigenvalues.

The paper proposes two variants. \textbf{INGC} (Algorithm~1) minimizes Eq.~\eqref{eq:ipe} by block coordinate descent, with a Hardmax that rounds $\Lift$ to a partition at each iteration. \textbf{SINGC} (Algorithm~2) drops the inner-product error, which the Grassmann term already controls on smooth signals, and rounds $\Lift$ once at the end.
 SINGC thus never uses the features. We keep it as the default, the simpler of the two variants (App.~\ref{app:ipe-instability}).

No official code was released, so we coded IPE from the paper (App.~\ref{app:ipe-instability}). We also extend IPE to the $\Delta$-Laplacians (App.~\ref{app:extensions}). IPE is one of the four methods whose hyperparameters we tune on the block model (App.~\ref{app:hp-sensitivity}).

\emph{Complexity.} $\mathcal{O}(\Nnodes \nclust^2)$ per iteration, as for FGC, plus one eigensolve for $U_k$. \emph{Implementation.} Native torch, GPU; ARPACK on CPU for $U_k$. \emph{Scalability.} Small; on PubMed at $r = 0.9$, the default run collapses $\Lift$ onto a few supernodes, and the pseudo-inverse in the log-determinant gradient no longer converges.

\section{Metric definitions}
\label{app:metric-definitions}
We give the formal definition of each metric of Sec.~\ref{sec:metrics}, grouped by the
same families. We write $\Pi = \Lift \Proj$ and $X_c = \Proj X$ throughout.

\subsection{Spectral fidelity}
\label{app:met-spectral}
The \textbf{Restricted Spectral Approximation} (RSA) of Loukas~\cite{loukas2019graph} is the worst-case reconstruction error over a subspace $R$, the span of the first $K$ eigenvectors of $\Lap$,
\begin{equation}
  \epsilon_{R, \Proj, \Lift} = \sup_{x \in R,\ \|x\|_\Lap = 1} \|x - \Pi x\|_\Lap ,
  \qquad \Pi = \Lift \Proj ,
  \label{eq:rsa}
\end{equation}
with $\|x\|_\Lap = \sqrt{x^\top \Lap x}$ the Laplacian semi-norm. The constant depends on $R$, $\Proj$, $\Lift$ and $\Lap$, hence on the choice of $\Delta$-Laplacian (Sec.~\ref{sec:metrics}). Since $\|x_c\|_{\Lapc} = \|\Pi x\|_\Lap$ for $x_c = \Proj x$ and $\Lapc = \Lift^\top \Lap \Lift$, the triangle inequality turns Eq.~\eqref{eq:rsa} into a bound on the energy of the signals of $R$,
\begin{equation}
  (1 - \epsilon)\, \|x\|_\Lap \;\le\; \|x_c\|_{\Lapc} \;\le\; (1 + \epsilon)\, \|x\|_\Lap,
  \qquad x \in R.
  \label{eq:rsa-bound}
\end{equation}
A small $\epsilon$ thus guarantees that the coarsening preserves the Laplacian energy of every signal of $R$. The previous implementations use a full eigendecomposition of $\Lap$~\cite{loukas2019graph, joly2024graph, joly2026taxonomy}. We compute the constant from the $K$ preserved eigenpairs only, by one sparse eigensolve, which scales to much larger graphs. With $K = 100$, this takes about $1$ second on Cora, $3$ on PubMed and $80$ on ogbn-arxiv, whatever the coarsening.

With $\lambda_i$ and $\tilde{\lambda}_i$ the $i$-th smallest nonzero eigenvalues of $\Lap$ and of $\Lapc$, the \textbf{relative eigenvalue error} (\texttt{mean\_REE}) and the \textbf{spectral distance} (\texttt{spectral\_distance}) are
\begin{equation}
  \mathrm{REE} = \frac{1}{m} \sum_{i=1}^{m} \frac{|\tilde{\lambda}_i - \lambda_i|}{\lambda_i},
  \qquad
  \mathrm{SD} = \Big( \sum_{i=1}^{\nclust - 1} (\lambda_i - \tilde{\lambda}_i)^2 \Big)^{1/2}.
  \label{eq:ree-sd}
\end{equation}
\texttt{mean\_REE} uses the $m$ smallest. \texttt{spectral\_distance} compares the whole spectrum of $\Lapc$ with the $\nclust - 1$ smallest nonzero eigenvalues of $\Lap$. The larger eigenvalues of $\Lap$ are left out.
The first is used across the coarsening literature~\cite{loukas2019graph, kumar2023featured, kataria2024ugc}. The second comes from Wilson and Zhu~\cite{wilson2008study}, and GBGC~\cite{xia2025gbgc} reports it.

\subsection{Operator reconstruction}
\label{app:met-operator}
These two metrics compare an operator of the graph with its approximation through $\Pi$. They use no features. The \textbf{reconstruction error} (\texttt{reconstruction\_error}) compares $\Lap$ with $\Pi^\top \Lap \Pi$, the Laplacian reduced to the coarsened graph and lifted back. The \textbf{Gromov-Wasserstein distortion} (\texttt{gromov\_wasserstein\_distortion}) of Taveras et al.~\cite{taveras2026gromov} compares a structure matrix $\Prop$ with its block-mean approximation $\Pi \Prop \Pi$, weighted by a node measure $\mu$. It is the Gromov-Wasserstein cost between the graph and its coarsened graph when the transport plan sends each node to its supernode. It has a closed form. The two metrics are
\begin{equation}
  \mathrm{RcE} = \frac{\| \Lap - \Pi^\top \Lap\, \Pi \|_F}{\|\Lap\|_F},
  \qquad
  D_{\mathrm{GW}} = \big\| (\Prop - \Pi \Prop \Pi) \odot (\mu \mu^\top)^{1/2} \big\|_F^2 .
  \label{eq:rce-gw}
\end{equation}
A \texttt{normalize} argument returns the raw squared norm for the first.

\subsection{Features}
\label{app:met-features}
These metrics compare the features $X$ with the coarsened features $X_c$.
The \textbf{feature RSA} (\texttt{rsa\_features}) applies the measure of the RSA (Eq.~\eqref{eq:rsa}) to each feature column $x_j$, as the relative error $\|x_j - \Pi x_j\|_\ast / \|x_j\|_\ast$, and needs no eigendecomposition. Its \texttt{norm} argument chooses $\|\cdot\|_\ast$. The default, $\|\cdot\|_\Lap$, measures the spectral fidelity of the features. The $\epsilon$-similarity of FGC~\cite{kumar2023featured}, reused by UGC~\cite{kataria2024ugc}, bounds the same quantity on the whole feature matrix, $\|X - \Pi X\|_\Lap \le \epsilon \|X\|_\Lap$; we report it per column. With $\|\cdot\|_2$, the measure becomes a reconstruction error of the features that no longer depends on $\Lap$. It is then below one, since $\Pi$ is an orthogonal projection.

The \textbf{Dirichlet energy error} (\texttt{dirichlet\_energy\_error}) compares the energy of the features with that of the coarsened features,
\begin{equation}
  \mathrm{DEE} = \frac{\big|\tr(X^\top \Lap X) - \tr(X_c^\top \Lapc X_c)\big|}{\tr(X^\top \Lap X)} .
  \label{eq:dee}
\end{equation}
A \texttt{mode} argument returns its logarithmic form $|\log(\tr(X^\top \Lap X) / \tr(X_c^\top \Lapc X_c))|$, as in IPE~\cite{cohen2026spectral}. The \textbf{hyperbolic error} (\texttt{hyperbolic\_error}) of Bravo-Hermsdorff and Gunderson~\cite{bravo2019unifying}, reused by UGC, FGC and IPE~\cite{kataria2024ugc, kumar2023featured, cohen2026spectral}, compares the residual $(\Lap - \Pi^\top \Lap \Pi) X$ with the two energies through a hyperbolic distance,
\begin{equation}
  \mathrm{HE} = \operatorname{arccosh}\!\left( 1 +
  \frac{\|(\Lap - \Pi^\top \Lap\, \Pi) X\|_F^2 \; \|X\|_F^2}
       {2\,\tr(X^\top \Lap X)\; \tr(X^\top \Pi^\top \Lap\, \Pi X)} \right).
  \label{eq:he}
\end{equation}
The next two metrics are close to the costs that two methods of the toolkit minimize.
The \textbf{inner-product error} (\texttt{inner\_product\_error}) of IPE~\cite{cohen2026spectral} compares the Gram matrices of the features in the $\Lap$ inner product, before and after coarsening. The \textbf{convolution-matching error} (\texttt{convolution\_matching\_error}) of ConvMatch~\cite{dickens2024graph} is the entrywise $\ell_1$ difference between one graph convolution and its lifted coarse counterpart,
\begin{equation}
  \mathrm{IPE} = \| X^\top \Lap X - X_c^\top (\Lift^\top \Lap \Lift) X_c \|_F^2,
  \qquad
  \mathrm{CME} = \| \Lift\, \Sc\, \Proj X - \Prop X \|_{1,1},
  \label{eq:ipe-cme}
\end{equation}
with $\Sc = \Proj \Prop \Lift$ for the pair of the coarsening, so that the lifted output is $\Pi \Prop \Pi X$. The objective of Dickens et al.\ mixes two pairs. The toolkit also computes it (\texttt{mode="dickens"}, App.~\ref{app:convmatch-pq}).

The \textbf{feature Wasserstein distance} (\texttt{wasserstein\_feature\_distance}) of OTCoarsening~\cite{ma2021unsupervised} is an entropic transport cost between the features of the nodes and those of the supernodes,
\begin{equation}
  W_\gamma(X, X_c) = \min_{T} \langle T, M \rangle - \gamma\, E(T),
  \label{eq:wfd}
\end{equation}
with $M_{ij} = \|X_i - X_{c,j}\|_2^2$ the cost of moving node $i$ to supernode $j$, $T$ a transport plan between uniform masses on the nodes and on the supernodes, and $E(T)$ its entropy. $W_\gamma$ ignores which node went to which supernode: it compares the features as \emph{distributions}. We solve it by Sinkhorn iterations, stopped once the supernode marginal is met within a tolerance.

\subsection{Inside the supernodes}
\label{app:met-supernodes}
The \textbf{maximum accuracy} (\texttt{max\_acc}) is an upper bound on the accuracy of a model that predicts one label per supernode, e.g.\ a classifier trained on the coarsened graph. At best, it labels correctly the nodes that have the majority label of their supernode,
\begin{equation}
  \mathrm{max\_acc} = \frac{1}{\Nnodes} \sum_{k=1}^{\nclust} \max_{c}\, |\{ i \in V_k : y_i = c \}| .
  \label{eq:maxacc}
\end{equation}
It needs the labels but no training, and gives a cheap estimate of how much a coarsening can hurt a node classification task.

The \textbf{partition density quality} (\texttt{partition\_density\_quality}) averages the granular-ball quality $q$ of GBGC~\cite{xia2025gbgc} (Eq.~\eqref{eq:gbgc}) over the supernodes of more than one node,
\begin{equation}
  \mathrm{PDQ} = \frac{1}{|\mathcal{K}|} \sum_{k \in \mathcal{K}} q(V_k),
  \qquad
  \mathcal{K} = \{ k : |V_k| > 1 \}.
  \label{eq:pdq}
\end{equation}
A high value means dense supernodes with many triangles, a low value sparse, chain-like clusters.

\subsection{Scoring the methods on Cora and PubMed}
\label{app:cross-scoring}
Each method coarsens Cora and PubMed at $r = 0.7$ with its default hyperparameters, its own operator (Table~\ref{tab:laplacian}) and seed 14. We then build $\Proj$ and $\Lift$ under the self-loop normalized Laplacian $\Lap = I_\Nnodes - \Prop$, with $\Prop$ the GCN propagation, and compute the RSA and the REE with the same $\Lap$. CME uses $\Prop$ itself, and \texttt{max\_acc} needs no operator. The scores of one column thus compare across methods. Table~\ref{tab:cross-scoring} reports the RSA ($K = 100$), the mean REE ($m = 100$), the CME and \texttt{max\_acc}.

\begin{table}[H]
\centering
\caption{Four metrics on each method at $r = 0.7$, under the self-loop normalized Laplacian. $r$ is the ratio each method actually reaches. CME is in thousands, \texttt{max\_acc} in percent. Lower is better for RSA, REE and CME, higher for \texttt{max\_acc}. The best score of each column is in bold. STPGC runs without labels, GBGC with its breadth-first initialization, IPE with its default variant SINGC. OOM: out of memory on a 48~GB GPU.}
\label{tab:cross-scoring}
\resizebox{\linewidth}{!}{%
\begin{tabular}{@{}lcccccccccc@{}}
\toprule
& \multicolumn{5}{c}{Cora} & \multicolumn{5}{c}{PubMed} \\
\cmidrule(lr){2-6}\cmidrule(lr){7-11}
Method & $r$ & RSA & REE & CME & \texttt{max\_acc} & $r$ & RSA & REE & CME & \texttt{max\_acc} \\
\midrule
\multicolumn{11}{@{}l}{\emph{Global clustering}} \\
Random & 0.71 & 7.14 & 10.03 & 58.7 & 48.4 & 0.71 & 4.43 & 12.59 & 37.2 & 61.8 \\
METIS & 0.72 & 1.95 & 3.20 & 37.9 & 86.0 & 0.79 & 2.49 & 3.66 & 28.7 & 84.3 \\
Graclus & 0.73 & 1.94 & 3.62 & 35.7 & 87.6 & 0.71 & 2.99 & 3.48 & 25.8 & 85.9 \\
Leiden & 0.70 & 1.58 & 3.40 & \textbf{29.2} & 89.3 & 0.70 & 1.59 & 4.54 & 22.3 & 87.7 \\
UGC & 0.71 & 7.01 & 8.92 & 59.6 & 49.2 & 0.71 & 5.26 & 12.67 & 36.5 & 61.5 \\
GW KGPC & 0.70 & 6.55 & 4.76 & 49.0 & 70.7 & 0.71 & 5.61 & 3.96 & 30.8 & 70.6 \\
\midrule
\multicolumn{11}{@{}l}{\emph{Representative selection}} \\
Kcenter & 0.70 & 1.56 & 2.38 & 29.5 & 90.1 & 0.70 & 2.43 & 0.97 & \textbf{21.0} & 88.7 \\
Herding & 0.70 & 1.93 & 2.65 & 32.4 & 89.7 & 0.70 & 3.93 & 2.64 & 24.1 & 88.5 \\
OTCoarsening & 0.65 & 1.75 & 2.00 & 34.3 & 88.4 & \multicolumn{5}{c}{OOM} \\
\midrule
\multicolumn{11}{@{}l}{\emph{Greedy contraction}} \\
Local Var. & 0.70 & \textbf{0.67} & \textbf{1.16} & 32.9 & \textbf{90.8} & 0.70 & \textbf{0.34} & \textbf{0.08} & 23.2 & \textbf{89.1} \\
ConvMatch & 0.70 & 2.49 & 2.25 & 33.7 & 90.3 & 0.70 & 4.83 & 1.70 & 23.4 & 88.4 \\
NOPE & 0.70 & 2.07 & 3.85 & 36.0 & 89.0 & 0.70 & 3.30 & 3.83 & 25.2 & 86.4 \\
GW GPC & 0.70 & 3.43 & 2.52 & 38.0 & 81.0 & 0.70 & 2.42 & 1.53 & 23.7 & 84.2 \\
STPGC & 0.70 & 3.14 & 4.11 & 41.3 & 83.6 & 0.70 & 3.26 & 5.05 & 25.5 & 86.2 \\
\midrule
\multicolumn{11}{@{}l}{\emph{Recursive splitting}} \\
SGWL & 0.70 & 2.26 & 2.61 & 37.0 & 85.8 & 0.70 & 2.62 & 1.54 & 25.5 & 83.4 \\
GBGC & 0.71 & 2.05 & 3.38 & 38.0 & 88.5 & 0.64 & 5.46 & 12.39 & 29.2 & 84.3 \\
\midrule
\multicolumn{11}{@{}l}{\emph{Dense optimization}} \\
FGC & 0.70 & 7.16 & 9.47 & 52.7 & 65.0 & 0.70 & 5.38 & 12.67 & 36.3 & 60.5 \\
IPE & 0.70 & 6.84 & 11.09 & 57.9 & 48.6 & 0.70 & 5.55 & 8.04 & 34.6 & 65.7 \\
\bottomrule
\end{tabular}}
\end{table}

Local variation has the lowest RSA and REE and the highest \texttt{max\_acc} on both graphs. For the RSA this is expected, since its local cost bounds it; note also that it builds its partition with the Laplacian used for scoring. Under CME, local variation ranks fourth on Cora and third on PubMed. The best CME goes to Leiden on Cora and to Kcenter on PubMed. ConvMatch, whose objective mixes two pairs where CME uses one (App.~\ref{app:convmatch-pq}), ranks only fifth and fourth. ConvMatch on PubMed shows the widest gap between metrics: fourth under CME and fifth under REE, it has a worse RSA than Random. At their default hyperparameters, UGC, FGC and IPE score close to Random under RSA, REE and CME, and well below the other methods under \texttt{max\_acc}. These defaults may misrank FGC and IPE: on the SBM, a grid search has raised their \texttt{max\_acc} from $0.50$ to $0.86$ and from $0.49$ to $0.84$, while UGC does not improve (App.~\ref{app:hp-sensitivity}). We did not tune them on Cora and PubMed. Note that the ratio is not met exactly by all methods: METIS returns fewer supernodes on PubMed ($r = 0.79$) and GBGC more ($r = 0.64$), which penalizes the first and favors the second.

\section{Method remarks and extensions}
\label{app:pseudocodes}

Reimplementing a published method involves choices that the paper does not always fix. The paper and the released code may disagree, a constant may be set only in a script, or a corner case may be left open. These choices can change the output. We make ours explicit in this appendix. Method by method, we state whether we followed the paper or the released code, and why. The details are mostly useful to a reader who wants to compare one of our numbers with another published implementation. We also describe the extensions we added to each method. Most of them are options. The few that change the default output of a method are marked with a $^\dagger$ in the two tables below. Table~\ref{tab:laplacian} (App.~\ref{app:laplacian-table}) records the operator each method uses internally to choose its partition. Table~\ref{tab:extensions} (App.~\ref{app:extensions}) summarizes the other extensions. Many of them come from one choice the toolkit leaves open: the Laplacian a coarsening is consistent with and thus the corresponding pair $\Lift$, $\Proj$. Both can change the intermediate costs of a method.

\subsection{Operator conventions per method}
\label{app:laplacian-table}
Section~\ref{sec:notation} separates two roles of the Laplacian. The \texttt{laplacian\_kind} passed to \texttt{build\_coarsening} fixes the Laplacian the lifting $\Lift$ is consistent with. The operator a method uses \emph{internally} to choose its partition is a separate choice, which Table~\ref{tab:laplacian} records for each method, with the alternatives from the original paper and those we add. Every method defaults to the operator of its original paper, with three exceptions. Local variation defaults to the self-loop normalized Laplacian of~\cite{joly2024graph, joly2026taxonomy, joly2026core}, the operator a graph convolution propagates with. Kcenter and Herding were designed without a graph, and we run them on two hops of graph convolution on the features. 

\begin{table}[H]
\centering
\caption{The operator each method uses to choose its partition. $\Prop$ is the GCN propagation $\hat{D}^{-1/2}(\Adj + I)\hat{D}^{-1/2}$. $^\dagger$ The three methods whose default differs from their original paper.}
\label{tab:laplacian}
\small
\begin{tabular}{@{}>{\raggedright\arraybackslash}p{0.16\linewidth}>{\raggedright\arraybackslash}p{0.24\linewidth}>{\raggedright\arraybackslash}p{0.24\linewidth}>{\raggedright\arraybackslash}p{0.26\linewidth}@{}}
\toprule
Method & Default in the toolkit & Alternatives from the original paper & Alternatives we add \\
\midrule
UGC & features and $\Adj$, mixed by $\alpha$ & features only & --- \\
GW GPC, GW KGPC & adjacency $\Adj$ & combinatorial $\Lap$, signless $\Lap$ & self-loop normalized $\Lap$, $\Prop$ \\
Kcenter & features after two hops, $\Prop^2 X$$^\dagger$ & raw features $X$ & any number of hops \\
Herding & features after two hops, $\Prop^2 X$$^\dagger$ & raw features $X$ & any number of hops \\
OTCoarsening & $\Prop$ and features & --- & --- \\
Local Variation & self-loop normalized $\Lap$$^\dagger$ & combinatorial $\Lap$ & normalized $\Lap$ \\
ConvMatch & $\Prop$ & --- & --- \\
NOPE & features (Euclidean) & --- & $\Prop$, combinatorial $\Lap$, self-loop normalized $\Lap$, adjacency \\
SGWL & adjacency $\Adj$ & --- & $\Prop$ \\
FGC & combinatorial $\Lap$ & --- & normalized $\Lap$, self-loop normalized $\Lap$ \\
IPE & combinatorial $\Lap$ & --- & normalized $\Lap$, self-loop normalized $\Lap$ \\
Random, METIS, Graclus, Leiden, GBGC, STPGC & \multicolumn{3}{l}{structural (adjacency only), no operator choice} \\
\bottomrule
\end{tabular}
\end{table}

\subsection{Summary of our extensions}
\label{app:extensions}
Table~\ref{tab:extensions} collects what our implementations add to their source, beyond the operators of Table~\ref{tab:laplacian}, which all implementations accept. Some additions are on by default: they reach a requested ratio, scale, make a run reproducible, or turn a method into a coarsener. The others are options. Without them, the implementation follows the source method. METIS runs as released and does not appear.

\begin{table}[H]
\centering
\caption{Additions and options of each implementation beyond its original paper, apart from the operators of Table~\ref{tab:laplacian}. $^\dagger$ Changes the output with respect to the source, as in Table~\ref{tab:laplacian}.}
\label{tab:extensions}
\small
\begin{tabular}{@{}l p{0.39\linewidth} p{0.39\linewidth}@{}}
\toprule
Method & On by default & Other options \\
\midrule
Random & -- & equal-size variant: a shuffled index list split into balanced chunks (the default draws cluster identifiers i.i.d.). \\
Graclus & seeding of the global torch random stream, from which the \texttt{torch-cluster} kernel samples its visit order; the kernel takes no seed argument. & -- \\
Leiden & log-scale binary search on the resolution to reach a requested ratio$^\dagger$, with a tolerance \texttt{n\_clusters\_tol} on the community count. & -- \\
UGC & projections of $X$ and of the sparse $\Adj$ computed separately, so the $\Nnodes \times (d + \Nnodes)$ augmented matrix of the released code is never formed densely. & -- \\
GW GPC, KGPC & -- & two $\Pi$ normalizations (paper: uniform); both coarse operators, $C^\top S C$ of the paper and $\Proj \Prop \Lift$, returned in \texttt{info}. The partition is the same for both. \\
Kcenter, Herding & recast as coarseners: pivots on the propagated features plus nearest-pivot assignment (the originals are coreset selectors)$^\dagger$; the assignment is computed by blocks, so the $\Nnodes \times \nclust$ distance matrix is never formed. & -- \\
OTCoarsening & hard partition taken from the row-wise argmax of $\Lift$$^\dagger$ (the paper works with the soft $\Lift$). & three treatments of an unassigned node through \texttt{uncovered}; three options to choose between the paper and the released code; the soft $\Lift$ returned optionally for graph classification (App.~\ref{app:otcoarsening-variants}). \\
Local Variation & vectorized edge cost; global reduction with the masses of the supernodes$^\dagger$; absorption pass to unstick stalls$^\dagger$ (App.~\ref{app:local-variation}). & per-level budget \texttt{n\_e\_percent}, proportional (default) or fixed; per-level reduction, the algorithm of Loukas with the combinatorial Laplacian; spectral norm of the neighborhood cost. \\
ConvMatch & -- & \texttt{pair}: the binary or the normalized pair for both the operator and the features, where the paper mixes them (App.~\ref{app:convmatch-pq}); \texttt{knn+edges} candidate set (paper: knn only). \\
NOPE & rebuild of the priority queue without its dead entries (\texttt{heap\_purge}). It bounds the memory and leaves all merges unchanged. & three merge variants (App.~\ref{app:nope-variants}). \\
STPGC & -- & label-free run with \texttt{label\_guide=False} (the paper method uses the train and validation labels); \texttt{coning\_once} and \texttt{incremental} follow Algorithms~1, 2 and~4 as printed, the default following the released code (App.~\ref{app:stpgc-variants}). \\
SGWL & \texttt{stop\_rule} splitting the largest part until the requested $\nclust$ is reached$^\dagger$ (the paper stops at a fixed recursion depth, the repository at a size threshold). & the repository's size threshold. \\
GBGC & the split loop stops at the requested number of supernodes$^\dagger$ (the paper has no ratio parameter). & \texttt{quality\_split\_factor} softening the accept rule; \texttt{init} and \texttt{accept} choosing between the two published variants (App.~\ref{app:gbgc-sqrtn}). \\
FGC & -- & both released solvers, through \texttt{variant}. \\
IPE & the sign of the Grassmann gradient opposite to the paper's$^\dagger$ (App.~\ref{app:ipe-instability}). & the sign of the paper. \\
\bottomrule
\end{tabular}
\end{table}

\subsection{OTCoarsening: three options, and the nodes the method leaves out}
\label{app:otcoarsening-variants}

The released code of Ma \& Chen~\cite{ma2021unsupervised} differs from their paper on several points. On four of them we follow the paper: the score uses the raw features $X$ (the code uses the encoder output $Z$), Eq.~3 keeps only the $\nclust$ selected columns, Eq.~6 fixes the number of Sinkhorn iterations, and the loss keeps the entropic term of Eq.~7. Three other points change a result, and we provide them as options (Table~\ref{tab:otcoarsening-switches}).

\begin{table}[H]
\centering
\small
\caption{The three OTCoarsening options, and the variant we ship as the default.}
\label{tab:otcoarsening-switches}
\begin{tabular}{@{}llll@{}}
\toprule
Option & Paper & Released code & Default \\
\midrule
\texttt{score\_square}    & $\sigma(\hat{\Adj} X W_\alpha)$ (Eq.~2) & $\sigma\big((\hat{\Adj} X W_\alpha)^2\big)$ & code \\
\texttt{cut}              & $\alpha_s$, the top-$\nclust$ values (Eq.~3) & $\mathrm{ReLU}(\alpha - \alpha_{(\nclust)} + 10^{-7})$ & paper \\
\texttt{zero\_row\_filter} & no filter (Eq.~4, Eq.~9) & \texttt{get\_nonzero\_rows} & paper \\
\bottomrule
\end{tabular}
\end{table}

\paragraph{The squared score.} The released code squares the pre-activation of Eq.~2 before the sigmoid. The square ranks nodes by $|\alpha|$ and discards the sign. On Cora over four seeds, the two variants are within the spread of the seeds, so we keep the code, the version behind the published numbers.

\paragraph{The transport rows.} The loss of Eq.~\eqref{eq:otcoarsening} transports the rows of $X$ onto those of $X_c$, all of them in the paper. The released \texttt{get\_nonzero\_rows} is meant to drop the all-zero rows first. It sorts the rows by their sum and keeps as many as there are nonzero rows, starting from the smallest, so it drops the rows of largest sum and keeps the zero ones. When no row is zero, it only reorders the rows. This is the case on Cora, where the results do not change, and we default to the paper.

\paragraph{The nodes the method leaves out.} The lifting $\Lift$ keeps the columns of $\hat{\Adj}$ at the $\nclust$ selected nodes (Eq.~3 of the paper), so it has the sparsity of $\Adj + I$. A node further than one hop from every selected node is then \emph{uncovered}: its row of $\Lift$ is zero. The paper targets graph classification, where such a node simply drops out of the pooling. A partition has to place it, so the completion is ours. The \texttt{uncovered} argument provides three options. \texttt{attach} gives each uncovered node the supernode of the nearest covered one. \texttt{isolate} makes each uncovered node a supernode of its own. \texttt{isolate\_recursive}, our default, coarsens the result again with \texttt{isolate}, level after level, as their \texttt{MultiLayerCoarsening} does, and keeps the level whose count comes closest to the request.

Table~\ref{tab:otcoarsening-uncovered} shows why a completion is needed. Left alone, the argmax sends all uncovered nodes to supernode $0$, since its zero row ties all columns. That supernode swallows $1836$ of the $2485$ nodes of Cora at $r = 0.9$. \texttt{attach} repairs the partition, but it returns fewer supernodes than asked. \texttt{isolate} returns more, since each uncovered node adds one. The recursive mode corrects the overshoot of \texttt{isolate}. It comes closer to the request than \texttt{attach} at each ratio, with $252$ supernodes for $249$ at $r = 0.9$, at the price of up to four levels of training. Its purity is higher than that of \texttt{attach} at $r = 0.5$ and $0.7$, and a bit lower at $r = 0.9$. We keep it as the default, since it follows the multi-level scheme of the authors.

\begin{table}[H]
\centering
\small
\caption{OTCoarsening on Cora ($\Nnodes = 2485$) under the shipped defaults. The argmax returns fewer supernodes than requested; \texttt{attach} keeps that count and repairs the partition, \texttt{isolate} overshoots the request, and \texttt{isolate\_recursive} stacks isolate levels and keeps the one closest to the request.}
\label{tab:otcoarsening-uncovered}
\resizebox{\linewidth}{!}{%
\begin{tabular}{@{}ccc cc cc cc ccc@{}}
\toprule
& & \multicolumn{3}{c}{argmax alone} & \multicolumn{2}{c}{\texttt{attach}} & \multicolumn{2}{c}{\texttt{isolate}} & \multicolumn{3}{c}{\texttt{isolate\_recursive}} \\
\cmidrule(lr){3-5} \cmidrule(lr){6-7} \cmidrule(lr){8-9} \cmidrule(lr){10-12}
$r$ & asked $\nclust$ & $\nclust$ & purity & largest & purity & largest & uncovered & $\nclust$ & levels & $\nclust$ & purity \\
\midrule
0.5 & 1243 &  987 & 0.795 & 418  & 0.901 & 59  & 417  & 1404 & 1 & 1404 & 0.930 \\
0.7 &  746 &  656 & 0.661 & 1025 & 0.862 & 44  & 1024 & 1680 & 2 &  830 & 0.883 \\
0.9 &  249 &  231 & 0.464 & 1836 & 0.794 & 178 & 1835 & 2066 & 4 &  252 & 0.780 \\
\bottomrule
\end{tabular}}
\end{table}

\subsection{Local variation: our generalizations and a vectorized edge cost}
\label{app:local-variation}
We extend the algorithm of Loukas~\cite{loukas2019graph} on four points, and vectorize its edge cost. We derive and evaluate these choices in a forthcoming work, and only state here what each one does and its default in the toolkit.

\paragraph{$\Delta$-Laplacian family.} The original cost is derived for the combinatorial Laplacian $\Lap = D - \Adj$. We replace it by the family $\Lap = \Delta \mathcal{L} \Delta$ of~\cite{joly2024graph, joly2026taxonomy}, which covers the combinatorial, normalized and self-loop normalized Laplacians. The RSA cost, the spectral basis and the lifting between levels are all expressed through the chosen $\Delta$. The default is the self-loop normalized Laplacian (Table~\ref{tab:laplacian}), with $K = 100$ preserved eigenvectors.

\paragraph{Reduction between levels.} Each level reduces the preserved basis before the next one, and the cost projects with that same reduction. The bound of Loukas thus holds for it. We offer two reductions, for each $\Delta$. The default, \texttt{reduction="global"}, averages each supernode with the \textbf{masses} of its nodes, i.e.\ their $\Delta_{ii}^{-2}$ summed. It is the only weighted average per level whose product over the levels is $\Lift^+$, the reduction every metric of the toolkit uses. The other, \texttt{"per\_level"}, takes the pseudoinverse of each level, an orthogonal projection at each level. With the combinatorial Laplacian, it is the algorithm of Loukas. Its reduction, the product of the per-level pseudoinverses, is returned as \texttt{P\_chain} and stored with the partition (App.~\ref{app:storage}).

\paragraph{Per-level budget.} The original algorithm sets no limit on how many nodes a level may remove: each level aims at the whole remaining reduction. Our implementation lets each level contract a fraction \texttt{n\_e\_percent} of its current nodes, $10\%$ by default. The descent thus contracts many edges in its first levels, and fewer and fewer as the graph shrinks. A fixed budget, with the same number of contractions at each level, is also available. The last level removes only the contractions still needed to reach the requested size.

\paragraph{Stalls and absorption.} Within a level, a node can be included in at most one contraction, so a hub merges with only one of its neighbors per level. On graphs with a heavy-tailed degree distribution (such as a star), a level then removes far fewer nodes than its budget, which we call a \textbf{stall}. When a level removes less than half of its budget, our implementation lets a free node join a cluster already formed. The join is accepted if the enlarged cluster costs at most twice the worst cost accepted at that level, up to ten nodes per cluster. This pass is on by default, and how much it helps depends on the graph. Typically it is triggered only for extreme coarsening ratios, and never on the SBM.

\paragraph{Vectorized edge cost.} The reference implementation scores each candidate edge separately, in a Python loop. For an edge, the cost has a simple closed form. It is the squared distance between the two nodes in the spectral embedding, i.e.\ their rows $a_i, a_j$ in the $K$ preserved eigenvectors, rescaled by their degrees $d_i, d_j$ and multiplied by a factor that depends on the two degrees and the edge weight. Intuitively, merging two nodes is cheap when the preserved eigenvectors barely tell them apart. On the combinatorial Laplacian, it is $\frac{d_i + d_j}{2}\, \| a_i - a_j \|_2^2$, the cost of the reference code, which the per-level reduction preserves at each level. Under the global reduction, this form holds at the first level only. The factor of the next levels also depends on the masses of the two supernodes. We then score the edges by blocks with array operations. The variation neighborhood family has no such closed form, and uses the general cost of Eq.~\eqref{eq:loukas-cost} with the Frobenius norm of the reference code. The spectral norm of Eq.~6 of~\cite{loukas2019graph} is an option (\texttt{cost\_norm}).

\subsection{ConvMatch: the pair $(\Proj, \Lift)$ inside the cost, and a vectorized exact cost}
\label{app:convmatch-pq}

ConvMatch~\cite{dickens2024graph} merges pairs of supernodes greedily. It scores a merge by how much it changes one graph convolution, computed on the coarsened graph and lifted back. We make three remarks on this cost. First, it uses one pair $(\Proj, \Lift)$ for the operator and another one for the features, and our implementation lets the user choose the pair. Second, once vectorized, its exact form runs about as fast as its bound. Third, adding the edges to the candidate pairs helps the normalized pair.

\paragraph{The objective of Dickens et al.}
Their cost (Eq.~8 of~\cite{dickens2024graph}) compares one graph convolution on the original graph with its lifted counterpart on the coarsened one,
\begin{equation}
    \label{eq:convmatch-cost}
    \| \Lift_b \, \Sc^{\mathrm{H}} \, \Proj_b X - \Prop X \|_{1,1},
\end{equation}
where $\Lift_b$ is the binary partition matrix and $\Proj_b X$ averages the features of each supernode. The coarse operator follows Huang et al.~\cite{huang2021scaling}, $\Sc^{\mathrm{H}} = \tilde{D}'^{-1/2} \Lift_b^\top (\Adj + I) \Lift_b \, \tilde{D}'^{-1/2}$, with $\tilde{D}' = \Lift_b^\top \tilde{D} \Lift_b$ the summed degrees of each supernode. Note that it keeps the internal edges and the size of each supernode as self-loops. We write it with an H to reserve $\Sc$ for $\Proj \Prop \Lift$. Merging $u$ and $v$ into $w$ only changes the coarse outputs of $u$, $v$ and of their neighbors. ConvMatch scores a merge by this exact change (their Eq.~9). A-ConvMatch replaces it by an upper bound (their Theorem~1), cheaper to evaluate. Both only score candidate pairs, the nearest neighbors of each node in an SGC embedding (App.~\ref{sec:methods:convmatch}).

\paragraph{The cost mixes two pairs.}
The operator $\Sc^{\mathrm{H}}$ is exactly $\Proj_u \Prop \Lift_u$ for the lifting $\Lift_u = \tilde{D}^{1/2} \Lift_b \tilde{D}'^{-1/2}$, whose columns have unit norm, and $\Proj_u = \Lift_u^\top = \Lift_u^{+}$. This pair rescales the columns of the normalized pair of the self-loop $\Delta$-Laplacian~\cite{joly2026taxonomy}, $\Lift_n = \tilde{D}^{1/2} \Lift_b D_c^{-1/2}$ with $D_c$ the degrees of $\Lift_b^\top \Adj \Lift_b$ plus one, and $\Proj_n = \Lift_n^{+}$. The two pairs share the projection $\Pi_n = \Lift_n \Proj_n = \Lift_u \Proj_u$. Their coarse operators are similar, $\Proj_n \Prop \Lift_n = \Lambda \, \Sc^{\mathrm{H}} \Lambda^{-1}$ with $\Lambda_{kk} = \big( (d_k + 1) / (d_k + s_k) \big)^{1/2}$, $d_k$ the summed degree and $s_k$ the size of supernode $k$, so $\Lambda$ is the identity on singletons. The only change is that $\Lift_u$ does not preserve the Laplacian: $\Lift_u^\top \Lap \Lift_u$ takes the supernode sizes as self-loops, so it is not the self-loop $\Delta$-Laplacian of $\Lift_b^\top \Adj \Lift_b$. The taxonomy~\cite{joly2026taxonomy} only studies pairs that preserve it. It therefore uses $\Lift_n$, and so does the toolkit.

The features, however, are reduced and lifted by the binary pair. The cost of Eq.~\eqref{eq:convmatch-cost} thus \textbf{mixes two pairs}: the operator comes from the normalized pair, while the features go through the binary one. The binary lift copies the coarse output to all nodes of a supernode, where $\Pi_n$ shares it according to $\tilde{d}_i^{1/2}$. In the view of~\cite{joly2026taxonomy}, \textbf{a single pair} defines both the operator $\Proj \Prop \Lift$ and the projection $\Pi = \Lift \Proj$, and the lifted output is $\Pi \Prop \Pi X$. Our implementation adds the two consistent pairs through \texttt{pair}, with the paper as the default:
\begin{equation*}
\begin{array}{lll}
    \texttt{dickens}    & \Lift_b \, \Sc^{\mathrm{H}} \, \Proj_b X, & \\[2pt]
    \texttt{normalized} & \Lift_u \, \Sc^{\mathrm{H}} \, \Proj_u X = \Lift_n (\Proj_n \Prop \Lift_n) \Proj_n X & = \Pi_n \Prop \Pi_n X, \\[2pt]
    \texttt{binary}     & \Lift_b \, (\Proj_b \Prop \Lift_b) \, \Proj_b X & = \Pi_b \Prop \Pi_b X.
\end{array}
\end{equation*}
Compared with the paper, \texttt{binary} changes only the operator, and \texttt{normalized} only the outer pair.

\paragraph{One form for the three pairs.} To derive the exact cost and its bound once for the three pairs, we write the output of a supernode $k$ at a given level, i.e.\ the row $k$ of $\Sc \Proj X$, as
\begin{equation}
    h_k = \alpha_k \sum_{l} M_{kl} z_l,
\end{equation}
where $l$ runs over the supernodes of the level, $M_{kl}$ is non-zero when $l$ is $k$ itself or one of its neighbors, and $z_l \in \mathbb{R}^d$ is the reduced feature of supernode $l$, rescaled according to the pair (Table~\ref{tab:convmatch-forms}).

\begin{table}[H]
\centering
\small
\caption{The three pairs of our ConvMatch implementation in the common form $h_k = \alpha_k \sum_l M_{kl} z_l$. $m_l$ is the mean feature of supernode $l$, $s_l$ its size, $d_l$ the summed degree of its nodes and $\tilde{d}'_l = d_l + s_l$. The lift from $w$ back to $u$ multiplies $h_w$ by the last column.}
\label{tab:convmatch-forms}
\begin{tabular}{@{}lllll@{}}
\toprule
\texttt{pair} & $M$ & $\alpha_k$ & $z_l$ & lift $w \to u$ \\
\midrule
\texttt{dickens}    & $\Lift_b^\top (\Adj + I) \Lift_b$ & $\tilde{d}'^{-1/2}_k$ & $m_l \, \tilde{d}'^{-1/2}_l$ & $1$ \\
\texttt{binary}     & $\Lift_b^\top \Prop \Lift_b$      & $1 / s_k$         & $m_l$ & $1$ \\
\texttt{normalized} & $\Lift_b^\top (\Adj + I) \Lift_b$ & $(d_k + 1)^{1/2} / \tilde{d}'_k$ & $\tilde{d}'^{-1}_l \sum_{p \in V_l} \tilde{d}_p^{1/2} x_p$ & $\big( (d_u + 1) / (d_w + 1) \big)^{1/2}$ \\
\bottomrule
\end{tabular}
\end{table}

\paragraph{Exact cost and bound for the three pairs.} Dickens et al.\ derive the exact cost (their Eq.~9) and its bound (their Theorem~1) for their pair only. We check that both extend to the two other pairs, so that each of the three runs as ConvMatch and as A-ConvMatch. The derivation only needs three facts, which hold for every pair when $u$ and $v$ merge into $w$. First, $M = \Lift_b^\top B \Lift_b$ for a fixed $B$, so the row and the column of $w$ are the sums of those of $u$ and $v$. Second, $\alpha_l$ only depends on the supernode $l$. Third, $z_w$ and $\alpha_w$ only depend on $u$ and $v$. Hence the output of a neighbor $l$ changes by
\begin{equation}
    \alpha_l \big( M_{lu} (z_w - z_u) + M_{lv} (z_w - z_v) \big).
\end{equation}
The exact cost sums the change of the outputs of $u$, $v$ and of their neighbors. For the bound, the triangle inequality splits each neighbor term in two, $\|z_w - z_u\|_1 \sum_{l} \alpha_l M_{lu} + \|z_w - z_v\|_1 \sum_{l} \alpha_l M_{lv}$, with equality when $u$ and $v$ share no neighbor. The two sums are cached per supernode, so the bound costs $\mathcal{O}(d)$ per pair. For \texttt{pair=dickens}, we recover Theorem~1.

\paragraph{A vectorized exact cost.} The released code scores the candidate pairs one by one, in a Python loop. Our implementation scores them with array operations. It gathers the neighbors of all pairs in one pass, puts the two terms of a neighbor shared by $u$ and $v$ in one norm, and sums. To bound the memory, the pairs go in blocks under a fixed memory budget. A test against a dense recomputation of all the coarse outputs checks that the cost is still Eq.~9, for the three pairs. It runs one to two orders of magnitude faster than our earlier loop over pairs, and about as fast as A-ConvMatch (Table~\ref{tab:runtime}), for about a third more peak GPU memory on Cora and PubMed and the same on ogbn-arxiv. The partitions of A-ConvMatch are close to the exact ones (Table~\ref{tab:convmatch-pairs}).

\paragraph{A second candidate set.} Our option \texttt{knn+edges} adds the edges of the current graph to the candidate pairs. It is our addition, unused by default.

\paragraph{Which pair works best.} Table~\ref{tab:convmatch-pairs} compares the three pairs. With the candidates of the paper, they are close in each column, although the two consistent pairs lower the RSA on PubMed. With the edges added, the normalized pair has the lowest CME and the lowest RSA in each column, for the bound and for the exact cost. At $r = 0.9$, it also gains up to four points of \texttt{max\_acc} over the pair of the paper, and the binary pair mostly sits in between. Interestingly, with the edges added, the pair of the paper (\texttt{dickens}) lowers its \texttt{max\_acc} and raises its CME at $r = 0.9$. The \texttt{normalized} pair lowers its CME and its RSA in each column. The exact cost gives the same ranking.

\begin{table}[H]
\centering
\small
\caption{ConvMatch under the three pairs, with the exact cost (ConvMatch) and the bound (A-ConvMatch), and with the candidates of the paper (kNN) or the edges added. CME is the objective of Dickens et al., Eq.~\eqref{eq:convmatch-cost}, in thousands (\texttt{mode="dickens"} of the metric); acc is \texttt{max\_acc} in \%. Best value of each block in bold.}
\label{tab:convmatch-pairs}
\resizebox{\linewidth}{!}{%
\begin{tabular}{@{}llcccccccccccc@{}}
\toprule
& & \multicolumn{6}{c}{Cora} & \multicolumn{6}{c}{PubMed} \\
\cmidrule(lr){3-8} \cmidrule(lr){9-14}
& & \multicolumn{3}{c}{$r = 0.7$} & \multicolumn{3}{c}{$r = 0.9$} & \multicolumn{3}{c}{$r = 0.7$} & \multicolumn{3}{c}{$r = 0.9$} \\
\cmidrule(lr){3-5} \cmidrule(lr){6-8} \cmidrule(lr){9-11} \cmidrule(lr){12-14}
Cand. & \texttt{pair} & CME & RSA & acc & CME & RSA & acc & CME & RSA & acc & CME & RSA & acc \\
\midrule
\multicolumn{14}{@{}l}{\emph{Exact cost (ConvMatch)}} \\
kNN & \texttt{dickens} & 36.2 & \textbf{2.14} & \textbf{90.5} & 53.0 & \textbf{2.61} & \textbf{86.4} & 28.2 & 4.83 & 88.5 & 36.2 & 4.01 & 84.0 \\
 & \texttt{binary} & 36.0 & 2.16 & 90.4 & 52.8 & 2.82 & 85.7 & 27.5 & \textbf{2.69} & \textbf{88.6} & 35.1 & 3.48 & 84.2 \\
 & \texttt{normalized} & \textbf{35.5} & 2.16 & 90.2 & \textbf{51.6} & 2.78 & 86.1 & \textbf{27.4} & \textbf{2.69} & 88.5 & \textbf{34.9} & \textbf{2.98} & \textbf{84.3} \\
\addlinespace
kNN & \texttt{dickens} & 34.6 & 3.25 & 90.6 & 56.1 & 2.58 & 84.8 & 28.9 & 2.15 & 88.5 & 40.6 & 2.04 & 81.7 \\
+ edges & \texttt{binary} & 32.4 & 3.19 & \textbf{91.2} & 53.5 & 2.77 & 84.3 & 27.5 & 2.79 & 88.8 & 37.7 & 2.28 & 84.0 \\
 & \texttt{normalized} & \textbf{30.4} & \textbf{1.65} & 90.8 & \textbf{47.8} & \textbf{2.11} & \textbf{86.3} & \textbf{26.1} & \textbf{1.32} & \textbf{89.0} & \textbf{34.2} & \textbf{1.93} & \textbf{85.7} \\
\midrule
\multicolumn{14}{@{}l}{\emph{Bound (A-ConvMatch)}} \\
kNN & \texttt{dickens} & 36.5 & 2.49 & 90.3 & 53.2 & 2.70 & \textbf{86.2} & 28.4 & 4.83 & 88.4 & 36.5 & 3.39 & 84.0 \\
 & \texttt{binary} & 36.0 & 2.49 & \textbf{90.4} & 52.9 & 2.60 & \textbf{86.2} & 27.6 & \textbf{2.97} & \textbf{88.6} & \textbf{35.1} & \textbf{3.19} & \textbf{84.3} \\
 & \texttt{normalized} & \textbf{35.6} & \textbf{2.14} & 90.2 & \textbf{52.0} & \textbf{2.57} & 86.1 & \textbf{27.5} & \textbf{2.97} & 88.5 & 35.2 & 4.31 & 84.2 \\
\addlinespace
kNN & \texttt{dickens} & 36.2 & 4.15 & 90.0 & 57.3 & 3.87 & 82.7 & 29.4 & 2.06 & 88.3 & 41.0 & 3.00 & 81.6 \\
+ edges & \texttt{binary} & 32.4 & 3.25 & \textbf{90.9} & 52.8 & 3.11 & 83.9 & 27.6 & 2.29 & 88.7 & 37.9 & 3.18 & 83.7 \\
 & \texttt{normalized} & \textbf{30.5} & \textbf{1.85} & 90.7 & \textbf{47.8} & \textbf{2.06} & \textbf{85.4} & \textbf{26.2} & \textbf{1.49} & \textbf{89.0} & \textbf{34.5} & \textbf{2.02} & \textbf{85.2} \\
\bottomrule
\end{tabular}}
\end{table}

\subsection{NOPE: two released routines, and a shortcut that differs from Eq.~1}
\label{app:nope-variants}

Bai et al.~\cite{bai2026rethinking2} define two costs, NOPE (their Eq.~1) and its closed-form approximation NOPE$^{*}$ (their Eq.~2). Their repository implements them as two merge routines, selected by a flag \texttt{relax}: \texttt{False}, the default, for NOPE and \texttt{True} for NOPE$^{*}$. The NOPE routine produces the numbers reported in the paper, but its cost differs from Eq.~1. Our implementation therefore provides three variants: \texttt{nope} (Eq.~1, our default), \texttt{nope\_repo} (the NOPE routine of the repository) and \texttt{nope\_star} (NOPE$^{*}$).

\paragraph{\texttt{nope}: Eq.~1 of the paper.}
This variant computes $\mathcal{I}_{pq}$ as stated in Eq.~1, vectorized over the neighborhood union $\Ncal_p \cup \Ncal_q$.

\paragraph{\texttt{nope\_repo}: the NOPE routine of the repository.}
The repository scores a candidate pair as $w_{pq} (c_p + c_q + \delta)$, with $w_{pq} = |v_p^c| |v_q^c| / (|v_p^c| + |v_q^c|)$ the size weight of Eq.~\eqref{eq:nope}. The sum $c_p + c_q + \delta$ is meant to compute $\sum_{k \in \Ncal_p \cup \Ncal_q} (s_{kp} - s_{kq})^2$, the sum of Eq.~\eqref{eq:nope}. Here $c_p = \sum_{k \in \Ncal_p} s_{kp}^2$ and $c_q = \sum_{k \in \Ncal_q} s_{kq}^2$ sum the squared similarities over each node's neighborhood, and $\delta = \sum_{k \in \Ncal_p \cap \Ncal_q} (s_{kp} - s_{kq})^2$ runs over the intersection only. This saves time. Since $c_p$ and $c_q$ depend on one node each, they are cached and reused at each candidate merge, and only the small intersection is evaluated per pair.

\emph{The two sums differ.} Equation~1 sums a single quantity, $(s_{kp} - s_{kq})^2$, over the neighborhood union. For a fixed summand $f(k)$, inclusion-exclusion is exact,
\[
    \sum_{k \in \Ncal_p \cup \Ncal_q} f(k) = \sum_{k \in \Ncal_p} f(k) + \sum_{k \in \Ncal_q} f(k) - \sum_{k \in \Ncal_p \cap \Ncal_q} f(k).
\]
Applied to $f(k) = (s_{kp} - s_{kq})^2$, it would recover Eq.~1 from two neighborhood sums and one intersection sum. But this summand couples \emph{both} endpoints: $\sum_{k \in \Ncal_p} (s_{kp} - s_{kq})^2$ depends on $q$ through $s_{kq}$, so it cannot be cached per node. A per-node cache can only store a self-term such as $c_p$, which is \emph{not} the summand of Eq.~1.

Split the union into $A = \Ncal_p \setminus \Ncal_q$, $B = \Ncal_q \setminus \Ncal_p$ and $C = \Ncal_p \cap \Ncal_q$, and expand $(s_{kp} - s_{kq})^2 = s_{kp}^2 - 2 s_{kp} s_{kq} + s_{kq}^2$. Table~\ref{tab:nope-error} compares the two costs region by region.
\begin{table}[H]
\centering
\caption{Per-node contribution to the pair cost, by region of the neighborhood union.}
\label{tab:nope-error}
\begin{tabular}{lll}
\toprule
region & Eq.~1 (\texttt{nope}) & repository (\texttt{nope\_repo}) \\
\midrule
$A$ (only $p$) & $s_{kp}^2 - 2 s_{kp} s_{kq} + s_{kq}^2$ & $s_{kp}^2$ \\
$B$ (only $q$) & $s_{kp}^2 - 2 s_{kp} s_{kq} + s_{kq}^2$ & $s_{kq}^2$ \\
$C$ (both)     & $s_{kp}^2 - 2 s_{kp} s_{kq} + s_{kq}^2$ & $2 s_{kp}^2 + 2 s_{kq}^2 - 2 s_{kp} s_{kq}$ \\
\bottomrule
\end{tabular}
\end{table}
On $A$ and $B$, the repository keeps a single squared similarity and drops the cross term $-2 s_{kp} s_{kq}$ with the other square. On $C$, it counts the squared similarities twice, since $c_p$, $c_q$ and $\delta$ each contribute them. The two errors do not cancel. For instance, two nodes with equal features, $x_p = x_q$, have $s_{kp} = s_{kq}$ for every $k$, so Eq.~\eqref{eq:nope} gives a null cost, the ideal merge. The repository gives them $c_p + c_q > 0$, the larger as $p$ and $q$ resemble their neighbors. On random graphs we measured a relative deviation of $24\%$ to $228\%$ between \texttt{nope\_repo} and Eq.~1.

\paragraph{\texttt{nope\_star}: the NOPE$^{*}$ routine of the repository.}
This routine implements the closed form of NOPE$^{*}$ (Eq.~\eqref{eq:nope-star}), which weighs $\|x_p - x_q\|^2$ by the supernode sizes and by $|\Ncal_r|$. It matches the paper.

\subsection{STPGC: labels in the three operations, and where we differ from the code}
\label{app:stpgc-variants}

\paragraph{Label-guided dominance (Sec.~3.4 of the paper).}
Labels enter all three operations of STPGC. In strong collapse, the neighbors of $u$ that share its label are checked first when looking for a dominator, which decides \emph{which} node absorbs $u$. In neighborhood coning, the same ordering decides which neighbor becomes the artificial dominator, hence the merge target. In edge collapse, a dominated edge whose two endpoints share a label is skipped, so intra-class edges survive by construction. This last one is a hard filter, where the other two only set an order.

As published, STPGC thus uses the train and validation labels while it coarsens. Two settings isolate this. The default of \texttt{stpgc} enables label-guided dominance with \texttt{label\_split="train\_val"}, the setup of Table~1 of the paper. \texttt{label\_guide=False} disables it, which makes the method purely topological. The \texttt{label\_split} argument also offers \texttt{train\_only}, which leaves out the validation labels. On Cora, the two settings never return the same partition.

\paragraph{The heterophilic DropEdge is out of scope.}
Sec.~4 of the paper applies a DropEdge to the \emph{coarsened} graph before GNN training, at a $10\%$ rate, prioritizing edges between supernodes of different labels. The released code implements it as a deterministic deletion of heterophilic edges. We leave it out. It rewires the coarsened graph, which our \texttt{(Partition, info)} output does not include, and the paper frames it as a stabilizer for GNN training. A DropEdge can be applied at training time instead.

\paragraph{Components without training nodes are kept.}
Their \texttt{find\_component} deletes connected components under ten nodes that contain no training node. A partition maps every node, so we keep them. Their edge and strong collapses also compare the labels of all nodes. Our implementation only uses the labels of the split chosen by \texttt{label\_split}, so its partitions differ from theirs.

\paragraph{A law for $\theta_1$.}
Their $\theta_1$ is a per-dataset constant: $15$ on Cora and CiteSeer, $25$ on DBLP, $50$ on ogbn-arxiv, $100$ on ogbn-products. We fit a law in the graph size to these five numbers, $\theta_1 = \min(100, \max(15, \Nnodes^{1/3}))$, which returns $15$, $15$, $26$, $55$ and $100$ on the same five graphs. The floor and the ceiling recover the published values on the smallest and the largest graphs. The aim of this law is to give a reasonable initial value of $\theta_1$ on a new graph, before any tuning. As in their code, we also raise $\theta_1$ by one whenever the approximate phase raises $\rho$, a step Algorithm~1 does not show.

\paragraph{Two speed-ups of the released code.} The released code differs from the pseudo-code on two points that change the runtime and little else. First, the code runs neighborhood coning once, when strong and edge collapses stall, then switches to the approximate phase, as described in Sec.~3.4. Algorithm~4 instead repeats it at each stall. Second, the code only re-examines what the previous operation changed. The pseudo-code re-examines every candidate when it goes from one operation to the next. Our implementation follows the code on both points by default. The \texttt{coning\_once} and \texttt{incremental} arguments switch back to the pseudo-code.

\subsection{SGWL: which entry point, and when to stop splitting}
\label{app:sgwl-entry-points}

Recall from App.~\ref{sec:methods:sgwl} that SGWL updates a soft transport plan $T$ between the nodes and $K$ parts by proximal steps (Eq.~\eqref{eq:sgwl}). Each step multiplies $T^{(n)}$ entrywise by the kernel $\exp(-M^{(n)}/\beta)$, then rescales it to the masses $\mu_s$ of the nodes and $\mu_t$ of the parts. Each node goes to the part that receives most of its mass in the final plan.

\paragraph{A single split or a recursive descent.} The repository of Xu et al.~\cite{xu2019scalable} hosts two generations of the method. \texttt{graph\_partition} cuts the graph once into $K$ parts with a single Gromov-Wasserstein transport plan. \texttt{recursive\_graph\_partition} cuts it recursively in two, the scalable method. We implement the recursive descent. The single split is still available by setting the number of parts per level, \texttt{cluster\_num}, to the requested $\nclust$. We do not benchmark it. On a three-community SBM its purity is far below that of the descent, and its dense $\Nnodes \times \nclust$ plan takes $57$\,GB on ogbn-arxiv at $r = 0.5$.

\paragraph{Each routine comes with its own $\beta$.} A step only acts through $M^{(n)}/\beta$, so $\beta$ depends on the scale of the adjacency in $M^{(n)}$. The released scripts call the single split with $\Adj_{\text{sym}}/\Nnodes$ and $\beta = 2 \times 10^{-7}$, and the recursive descent with $\Adj_{\text{sym}}$ and $\beta = 0.15$. We take the second pair.

\paragraph{Two places where we follow the released code.} Algorithm~4 of the paper assigns a node to the part $\arg\max_j T_{ij}$. The released \texttt{node\_cluster\_assignment} first divides each column of $T$ by $\mu_{t,j}$, so a node goes to the part that receives the largest mass \emph{relative to the mass of that part}. Section~3.1 defines the node-prior cost as $c_{ij} = |\mu_{s,i} - \mu_{t,j}|$, and the released \texttt{node\_distribution\_similarity} squares it. We follow the code on both. On the SBM, the two versions of the assignment give the same number of supernodes and the same purity.

\paragraph{A stopping rule that targets a coarsening ratio.} The paper stops after a fixed number of recursions $R$, i.e.\ $2^R$ parts. The repository stops splitting a part once it has at most \texttt{max\_node\_num} nodes. Neither targets a coarsening ratio. We provide two rules through \texttt{stop\_rule}. The \texttt{size} rule reproduces the repository. The \texttt{count} rule, our default, always splits the largest part and stops once $\nclust$ parts exist. The size rule overshoots the requested count, since a part typically ends well below the threshold: cutting a part of $12$ under a threshold of $9$ yields parts of $7$ and $5$ (Table~\ref{tab:sgwl-stop}). The count rule can only miss from below, when a part is too small to cut. Both rules run about the same number of proximal steps.

\begin{table}[H]
\centering
\small
\caption{SGWL on Cora ($\Nnodes = 2485$) under the two stopping rules. The size rule is the one of the released code.}
\label{tab:sgwl-stop}
\begin{tabular}{@{}cccc@{}}
\toprule
$r$ & asked $\nclust$ & $\nclust$, \texttt{size} (repository) & $\nclust$, \texttt{count} (default) \\
\midrule
0.5 & 1243 & 1281 & 1243 \\
0.7 &  746 & 1057 &  746 \\
0.9 &  249 &  407 &  249 \\
\bottomrule
\end{tabular}
\end{table}

\subsection{GBGC: the $\sqrt{\Nnodes}$ target, and three options}
\label{app:gbgc-sqrtn}

Two papers of the same group define the method. GBGC, by Xia et al.~\cite{xia2025gbgc}, covers the graph with about $\sqrt{\Nnodes}$ granular-balls grown by breadth-first search, then splits them while a structural quality improves (App.~\ref{sec:methods:gbgc}). A later work~\cite{wang2026efficient} keeps this splitting scheme and changes three of its ingredients. We catalogue the first paper, and provide the changes of the second as options.

\paragraph{The initial ball count is a floor, and it is not $\sqrt{\Nnodes}$.} Eq.~3 of~\cite{xia2025gbgc} targets $\sqrt{\Nnodes}$ initial balls. Each ball grows by breadth-first search from the node of highest degree not yet covered, and stops before it exceeds $\sqrt{\Nnodes}$ nodes (Algorithm~2 of their appendix; Eq.~4 stops on the size of a layer instead). If every ball reached $\sqrt{\Nnodes}$ nodes, there would be $\sqrt{\Nnodes}$ balls. On a real graph, many balls stop early, since their neighbors already belong to earlier balls. On Cora ($\Nnodes = 2485$, $\lceil\sqrt{\Nnodes}\rceil = 50$), we obtain $401$ initial balls of about $6$ nodes, instead of $50$ balls of $50$ nodes. METIS, used by the later work, is asked for $\sqrt{\Nnodes}$ balanced parts and returns exactly that number. A split only raises the count. The initial count is thus a floor. No smaller count can be requested. At the other end, the splits stop on their own at $720$ balls. With the breadth-first initialization, GBGC thus only reaches the \textbf{feasible range} $r \in [0.710, 0.839]$ on Cora.

\paragraph{Three options.} Table~\ref{tab:gbgc-switches} lists the three changes of~\cite{wang2026efficient}, with the GBGC paper as the default. Our implementation also provides \texttt{quality\_split\_factor}, a factor on the right-hand side of Eq.~8 of~\cite{xia2025gbgc}, equal to $1$ in the paper.

\begin{table}[H]
\centering
\small
\caption{The three GBGC options. A ball $B$ is split into $B_1$ and $B_2$ when the criterion is met. The default is the paper we catalogue.}
\label{tab:gbgc-switches}
\resizebox{\linewidth}{!}{%
\begin{tabular}{@{}llll@{}}
\toprule
Option & \cite{xia2025gbgc} & \cite{wang2026efficient} & Default \\
\midrule
\texttt{init}   & BFS from the highest-degree centers (Eq.~4) & METIS asked for $\sqrt{\Nnodes}$ parts & BFS \\
\texttt{accept} & $q(B_1) + q(B_2) > q(B)$ (Eq.~8) & $\big(q(B_1) + q(B_2)\big)/2 > q(B)$ & sum \\
\texttt{use\_transitivity} & $q = E_B/\Nnodes_B + \text{transitivity}$ (Eq.~6) & $q = E_B/\Nnodes_B$ alone & true \\
\bottomrule
\end{tabular}}
\end{table}

Asking $\nclust = 249$ on Cora, the breadth-first initialization still gives its $401$ balls under both criteria. With METIS, the sum of the paper gives exactly $249$, and the mean returns $62$, since it accepts only $12$ splits. Only METIS with the sum reaches the request, and we report it as a second version, \texttt{gbgc\_metis} (App.~\ref{sec:methods:gbgc}).

\subsection{FGC: two released solvers}
\label{app:fgc-code}

The paper of Kumar et al.~\cite{kumar2023featured} presents FGC with and without a sparsity regularizer $h(L_c) = \|L_c\|_F^2$ (its Sec.~4.1). Their repository implements the two as separate notebooks, which also differ in normalization, initialization and step. Our two variants replicate these two notebooks, through a \texttt{variant} argument.

\paragraph{Metric solver (default).}
The notebook \texttt{FGC\_experiment.ipynb} produces the paper's REE, DE and HE tables. It has no sparsity regularizer ($\beta = 0$), normalizes the coarse features by their $\ell_2$ norm, initializes the lifting matrix as a dense Gaussian, and takes a fixed gradient step of $10^{-5}$ over ten outer iterations of one hundred inner steps. We use it as the default (\texttt{variant="metrics"}). The notebook and the appendix of the paper agree on the three weights of Eq.~\eqref{eq:fgc} for the main Cora run, $500$, $500$ and $F/2 = 716.5$ with $F$ the number of features, but not on which weight takes which value. Both put $\alpha = 500$ on the feature term. The notebook puts $F/2$ on $\gamma$, the weight of the log-determinant, and $500$ on $\lambda$, the weight of the group-sparsity penalty. The appendix of the paper reports $\gamma = 500$ and $\lambda = F/2$. \texttt{weights\_source} provides both settings, and defaults to the notebook.

\paragraph{Sparsity solver.}
The notebook \texttt{Node\_classification\_FGC\_with\_sparsity.ipynb} targets node classification. It adds the sparsity regularizer $\beta \| L_c \|_F^2$, normalizes by the $\ell_1$ norm, initializes sparsely, and takes a single inner step. We provide it as \texttt{variant="sparsity"}.

\paragraph{Step size.}
Algorithm~1 of the paper uses the true Lipschitz constant $L = \max(L_1, \dots, L_4)$ of the four gradient terms of Eq.~11 of~\cite{kumar2023featured}. Neither notebook computes it. The metric solver uses a fixed step of $10^{-5}$. The sparsity notebook sets a proxy $L = 1/\nclust$ and steps by $1/L$. Its effective step is thus $\nclust$. Our implementation uses these two steps for the two variants. Which weights and step to change on the SBM is discussed in App.~\ref{app:hp-sensitivity}.

\paragraph{Soft versus hard lifting matrix.}
The paper's metrics are computed on the continuous lifting matrix, which is not a partition. On Cora at $r = 0.7$, we reproduce a REE of $0.038$ on that matrix, against the $0.058$ reported. The hard partition given by its row-wise argmax gives $0.549$. Our implementation always returns the hard partition. The continuous matrix is stored in \texttt{info}, behind a flag that warns against using it as a coarsening.

\subsection{IPE: a sign in the gradient, and the default hyperparameters of the paper}
\label{app:ipe-instability}

No official code was released for IPE, so our implementation follows the gradients of Appendix~A.4 of Cohen \& Talmon~\cite{cohen2026spectral}.

\paragraph{A sign in the Grassmann gradient.}
The sign of the Grassmann gradient printed in Eqs.~16 and~17 of~\cite{cohen2026spectral} differs from the derivative of the term. The term enters the objective as $-\beta \tr(U_k U_k^\top \Lift \Lift^\top)$, so its gradient is $-2 \beta\, U_k U_k^\top \Lift$, while the paper prints $+2 \beta\, U_k U_k^\top \Lift$. With the printed sign, a descent step pushes the column space of $\Lift$ away from $U_k$. We provide both signs through \texttt{correct\_grassmann\_sign}, and default to the sign of the derivative.

\paragraph{IPE at the default hyperparameters of the paper.}
INGC rounds $\Lift$ to a partition at each outer iteration. A column left empty makes $\Lift^\top \Lap \Lift + J$ singular, and the log-determinant then drives the collapse further. On Cora at $r = 0.9$, INGC ends with $3$ supernodes for $249$ requested. SINGC rounds once, at the end, and returns the requested count, but no better than random on the SBM and on Cora. On PubMed at $r = 0.9$, SINGC fails and INGC matches random. Tuning on the SBM lifts SINGC from random to a purity of $0.836$ (App.~\ref{app:hp-sensitivity}). A tuning per dataset would thus likely help. It is however costly, with three weights, a step and two iteration counts to set. We keep SINGC as the default, the simpler of the two variants.

\section{Datasets and preprocessing}
\label{app:datasets}

The toolkit proposes loaders for ten node-classification graphs~:  three homophilic citation networks (Cora,
CiteSeer, PubMed), four heterophilic graphs (chameleon, squirrel, Amazon-ratings,
Roman-empire), and three large-scale graphs (ogbn-arxiv, ogbn-products, Reddit). We load them
with the standard PyG and OGB loaders.

Two preprocessing steps make the inputs match the assumptions of the methods (Sec.~\ref{sec:design}). The spectral costs are written for a Laplacian that is symmetric by construction, which also makes the eigensolvers faster on it, and Graclus requires a graph without isolated node.
First, we strip self-loops and symmetrize any directed input, since \texttt{pymetis} and the
edge-contraction methods assume an undirected graph. Only chameleon, squirrel and ogbn-arxiv
ship directed. Chameleon and squirrel also ship with self-loops. Second, we restrict a graph to
its largest connected component, since the cost of local variation is built on a single null eigenvector
of $\Lap$, i.e. a connected graph. We apply both steps to all datasets. Only Cora, CiteSeer and
ogbn-products have several components. CiteSeer loses about a third of its nodes. PubMed, chameleon, squirrel,
Amazon-ratings, Roman-empire, ogbn-arxiv and Reddit already form a single connected component,
and the step leaves them unchanged. Table~\ref{tab:datasets} summarizes the drop per dataset.

\begin{table}[H]
\centering
\caption{The ten datasets after preprocessing. \emph{Symmetrized}: the input graph is directed.
\emph{Before LCC}: number of nodes before we keep the largest connected component, for the three
graphs that are not connected.}
\label{tab:datasets}
\begin{tabular}{llrrrcr}
\toprule
Dataset & Type & Nodes & Feat. & Cls. & Symmetrized & Before LCC \\
\midrule
Cora           & homophilic   & 2,485     & 1,433 & 7  &            & 2,708 \\
CiteSeer       & homophilic   & 2,120     & 3,703 & 6  &            & 3,327 \\
PubMed         & homophilic   & 19,717    & 500   & 3  &            & -- \\
chameleon      & heterophilic & 2,277     & 2,325 & 5  & \checkmark & -- \\
squirrel       & heterophilic & 5,201     & 2,089 & 5  & \checkmark & -- \\
Amazon-ratings & heterophilic & 24,492    & 300   & 5  &            & -- \\
Roman-empire   & heterophilic & 22,662    & 300   & 18 &            & -- \\
ogbn-arxiv     & large-scale  & 169,343   & 128   & 40 & \checkmark & -- \\
ogbn-products  & large-scale  & 2,385,902 & 100   & 47 &            & 2,449,029 \\
Reddit         & large-scale  & 232,965   & 602   & 41 &            & -- \\
\bottomrule
\end{tabular}
\end{table}

\section{Experiments on synthetic graphs}
\label{app:experimental}

\subsection{The SBM at other ratios}
\label{app:sbm-ratios}

\begin{figure}[H]
\centering
\includegraphics[width=0.9\linewidth]{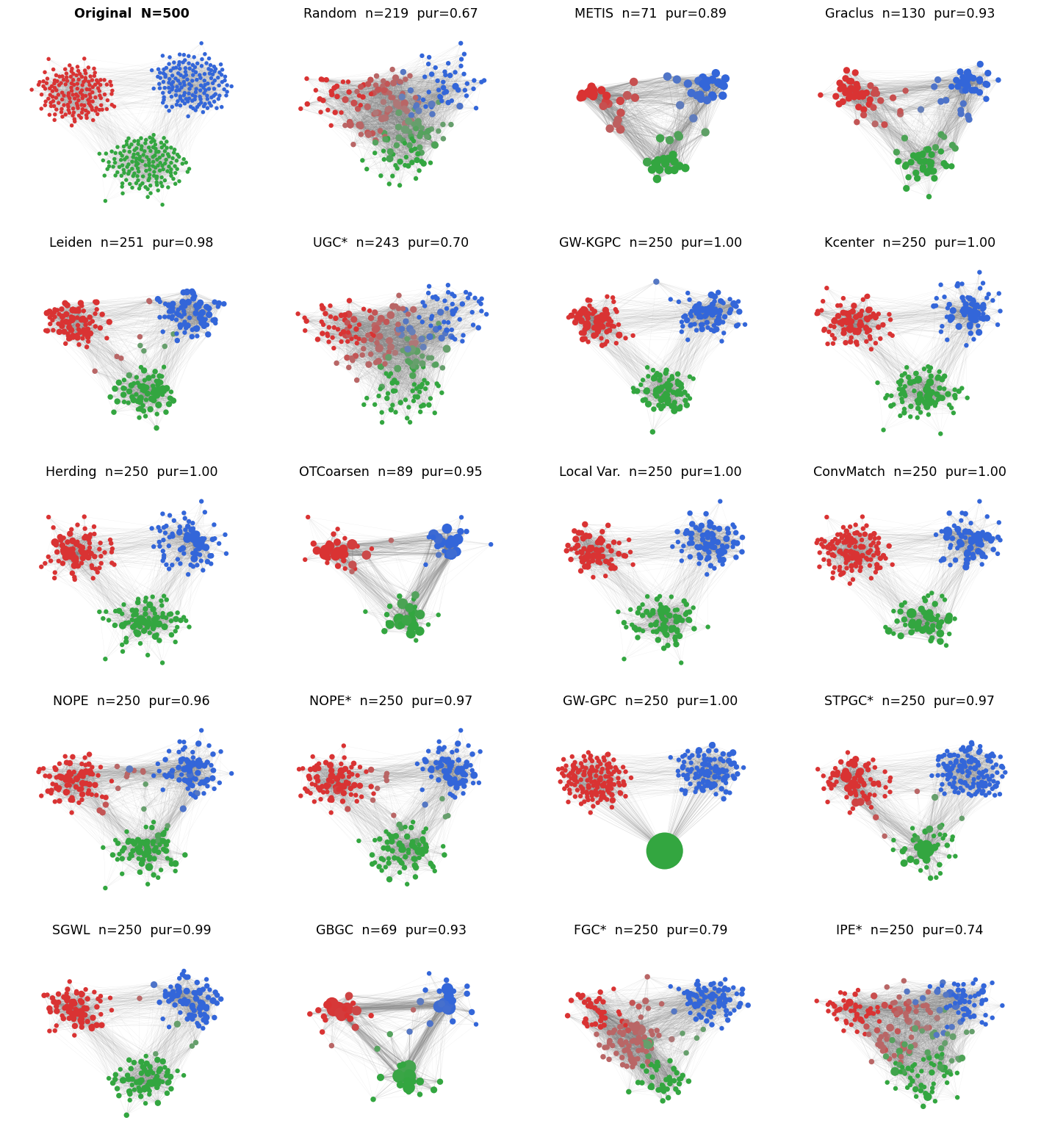}
\caption{Coarsened graph per method on the three-community SBM, at ratio $r = 0.5$.}
\label{fig:sbm-r05}
\end{figure}

\begin{figure}[H]
\centering
\includegraphics[width=0.9\linewidth]{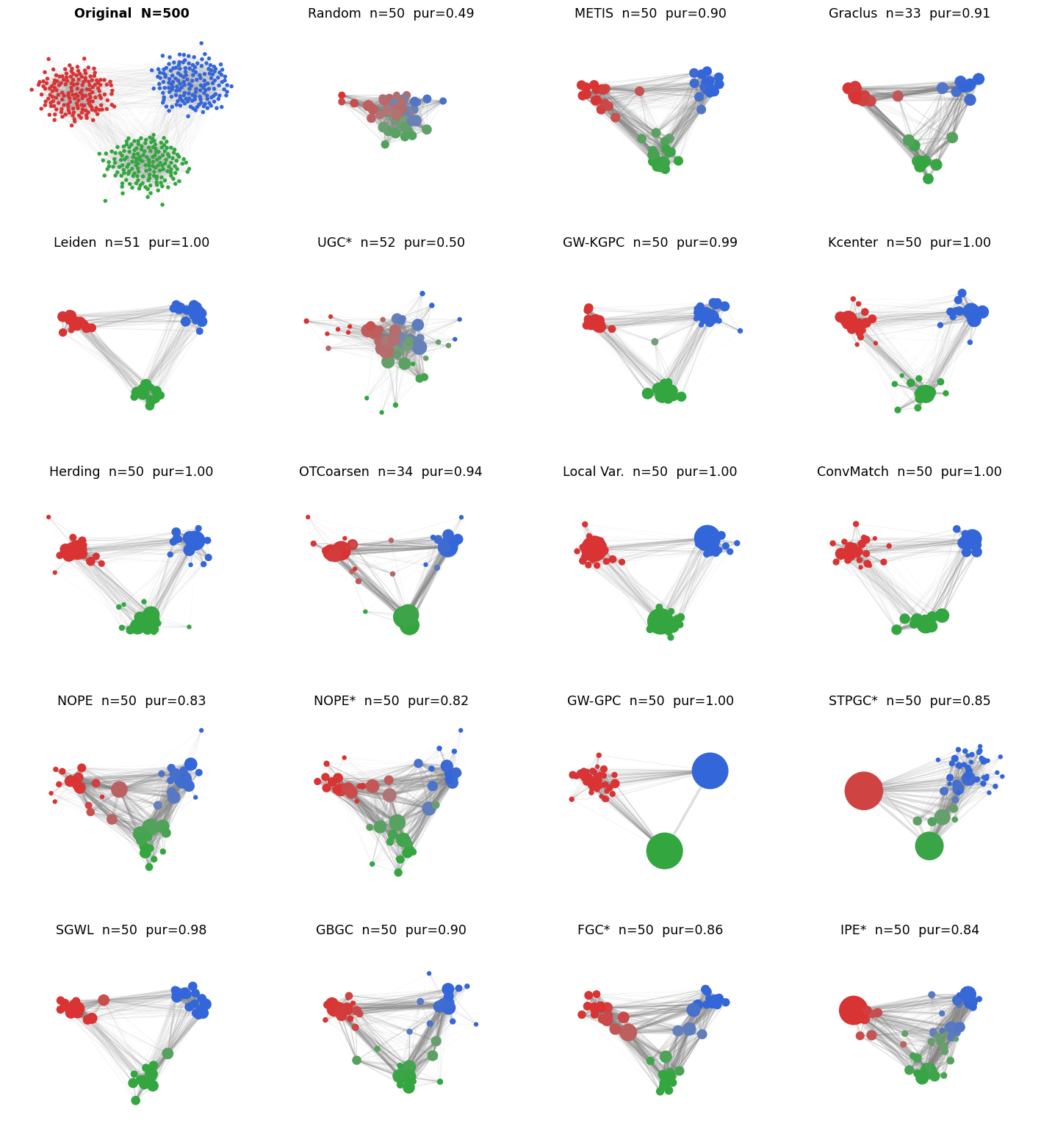}
\caption{Coarsened graph per method on the three-community SBM, at ratio $r = 0.9$.}
\label{fig:sbm-r09}
\end{figure}

Figure~\ref{fig:sbm} in the main text shows the coarsened graph of each method at the
ratio $r = 0.7$. Figures~\ref{fig:sbm-r05} and~\ref{fig:sbm-r09} show the
same panel at $r = 0.5$ and $r = 0.9$. The observations of Sec.~\ref{sec:sbm} hold across
ratios. The methods that preserve the communities at $0.7$ preserve them at each ratio. The concentrated
methods coarsen one community first. The gap widens at $0.9$, where fewer supernodes leave
less room to respect the three communities at once.

\subsection{Hyperparameter tuning}
\label{app:hp-sensitivity}

Four methods of Fig.~\ref{fig:sbm}, marked with *, run with hyperparameters tuned on the SBM:
UGC, STPGC, FGC and IPE. At their paper defaults, these four are well below the others. The tuning is done at $r = 0.9$, and the three SBM figures reuse the tuned configuration at each ratio. 

This subsection measures how much of this gap tuning their hyperparameters closes. 
The setup is still an SBM with three communities,
$\Nnodes = 500$ nodes, and coarsening ratio $r = 0.9$ (so $\nclust = 50$ supernodes). Node
features are drawn as $X \sim \mathcal{N}(0, \Lap^\dagger)$, as in Sec.~\ref{sec:sbm}.
Quality is the purity \texttt{max\_acc} of Sec.~\ref{sec:metrics}, with the communities as
labels: the fraction of nodes that belong to the majority community of their supernode. It
equals $1.0$ when every supernode lies inside one community. 

Table~\ref{tab:hp-sensitivity}
compares each method's default hyperparameters, usually the source paper's, against a grid
search. We grid the methods whose default scores below $0.9$, apart from NOPE ($0.83$). At their defaults, Leiden, Kcenter, Herding, local variation, ConvMatch
and GW GPC score $1.0$, and METIS, Graclus, GW KGPC, OTCoarsening, SGWL and GBGC lie
between $0.90$ and $0.99$.

\begin{table}[H]
\centering
\small
\caption{Purity \texttt{max\_acc} on the SBM ($r = 0.9$) under default versus tuned
hyperparameters. The tuned column is the best configuration of the grid that still returns the requested
$50$ supernodes.}
\label{tab:hp-sensitivity}
\begin{tabular}{lccl}
\toprule
Method & Default & Tuned & Key hyperparameter \\
\midrule
UGC           & 0.50 & 0.50 & none (insensitive) \\
STPGC         & 0.76 & 0.85 & $\theta_2$ raised from $1\%$ to $5\%$ of $\Nnodes$ \\
FGC           & 0.50 & 0.86 & learning rate $0.02$; $\alpha$, $\lambda$ lowered to $1$ and $10$, $\gamma$ raised to $500$; $200$ iterations \\
IPE           & 0.49 & 0.84 & Grassmann rank $50$, learning rate $10^{-3}$, $5000$ iterations \\
\bottomrule
\end{tabular}
\end{table}

\begin{table}[H]
\centering
\small
\caption{Grids behind Table~\ref{tab:hp-sensitivity}. $\nclust = 50$ is the requested number of supernodes.}
\label{tab:hp-grids}
\begin{tabular}{@{}lrp{10cm}@{}}
\toprule
Method & Configs & Grid \\
\midrule
UGC & 12 & adjacency in the hash: yes, no; projectors $\in \{100, 500, 2000\}$; distribution: uniform, normal \\
STPGC & 24 & $\theta_1 \in \{2, 3, 5, 8, 10, 15, 20, 30\}$, $\theta_2 / \Nnodes \in \{0.5\%, 1\%, 5\%\}$ \\
FGC & 1152 & solver: metrics, sparsity; learning rate $\in \{10^{-5}, 10^{-3}, 1/\nclust\}$ (metrics only); $\alpha \in \{1, 10, 100, 500\}$, $\gamma \in \{5, 100, 500\}$, $\lambda \in \{0.5, 1, 2, 5, 10, 50, 100, 500\}$; iterations $\in \{10, 50, 200\}$ \\
IPE & 432 & variant: SINGC, INGC; learning rate $\in \{10^{-4}, 10^{-3}, 1/\nclust\}$; $\alpha \in \{0.1, 1, 10, 100\}$, $\lambda \in \{0.1, 1, 100\}$; Grassmann rank $\in \{10, 50, 100\}$; iterations $\in \{2000, 5000\}$ for SINGC, $\{20, 5000\}$ outer steps for INGC \\
\bottomrule
\end{tabular}
\end{table}

UGC does not improve. None of its $12$ configurations beats its default, at $0.50$.
Neither the number of projectors nor the adjacency block changes that.

The best setting for STPGC keeps the default $\theta_1 = 15$ and raises $\theta_2$, the number of removed
nodes below which a round of the approximate phase increments $\rho$
(App.~\ref{sec:methods:stpgc}), from $1\%$ to $5\%$ of $\Nnodes$. It scores $0.85$. This may be
due to $\rho$ growing sooner, which on this graph keeps the merges inside the communities for
longer. The gain over the default is $0.09$. STPGC runs label-free here: its default uses the
labels, and on the SBM the labels are the planted communities.

FGC gains the most from tuning, over a grid of $1152$ configurations. Its default scores
$0.50$ and its best setting $0.86$: a learning rate of $0.02$ in place of $10^{-5}$, the
reconstruction weight $\alpha$ lowered from $500$ to $1$, the group-sparsity weight
$\lambda$ from $500$ to $10$, the log-det weight $\gamma$ raised from $F/2 = 5$ to $500$, over $200$ outer
iterations. The learning rate accounts for most of that gain. 
With the $10$ iterations of the authors' notebook, the same setting still scores $0.82$.

IPE scores $1.0$ over its grid of $432$ configurations, with a Grassmann rank of $10$, but gives only $5$ supernodes of the $50$ requested.
Among the configurations that return the
requested count, the best has $0.84$ purity, at rank $50$ with a learning rate of $10^{-3}$ over $5000$
iterations of SINGC. At the published iteration count it is $0.78$. FGC and IPE, the two dense
optimization methods, are the hardest to tune: each has several weights, a learning rate and
an iteration count, and their paper defaults score at the level of the random baseline on this graph.

\begin{figure}[H]
\centering
\includegraphics[width=0.8\linewidth]{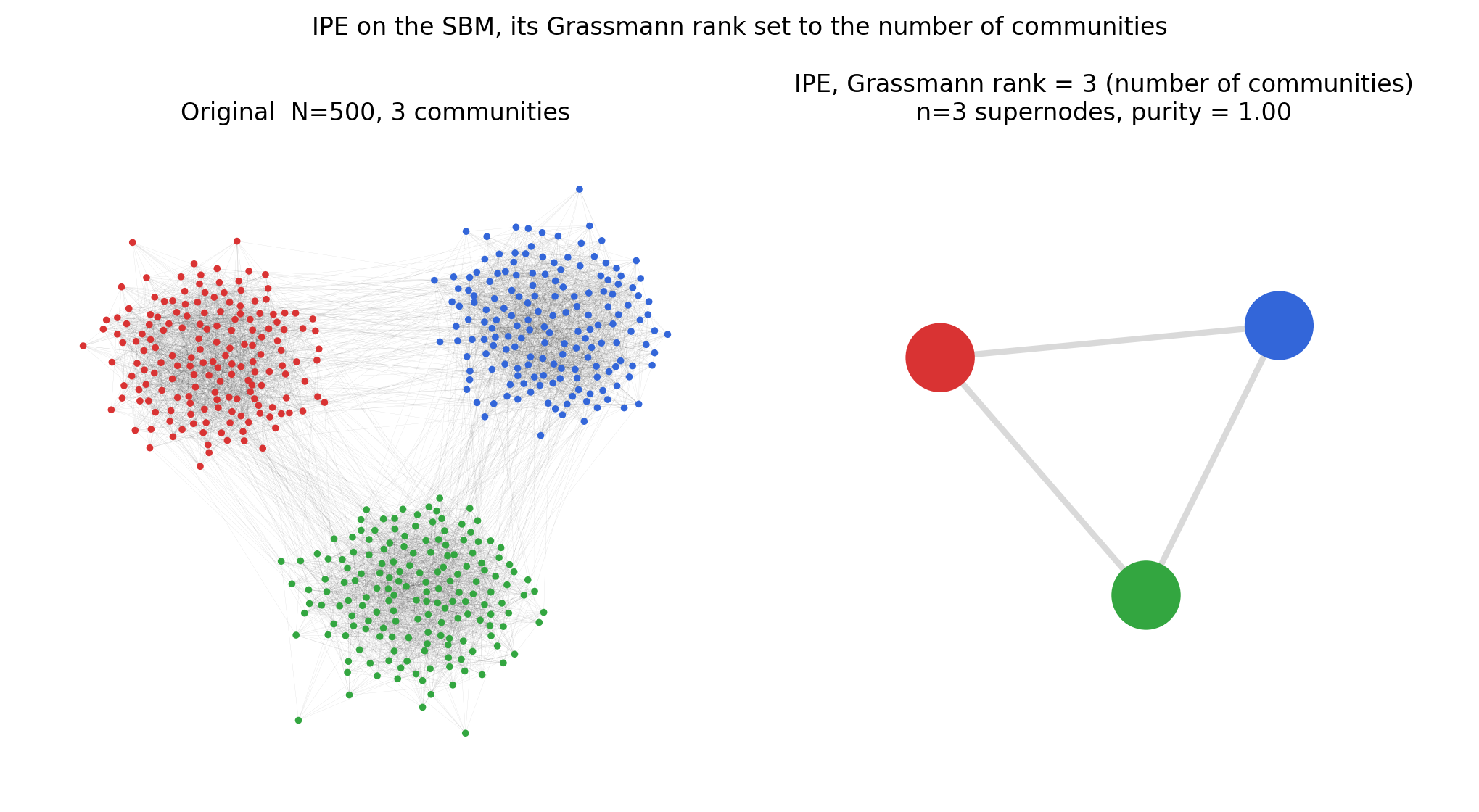}
\caption{IPE on the three-community SBM, with the Grassmann rank set to $3$ and the best configuration of Table~\ref{tab:hp-grids}. It returns three community-pure supernodes on all seeds.}
\label{fig:ipe-k}
\end{figure}

\paragraph{IPE and the Grassmann rank.}
\label{app:sbm-ipe}
IPE aligns the coarsened graph with as many leading eigenvectors of $\Lap$ as its Grassmann rank,
which by default equals the requested number of supernodes. With the rank set to the number of
communities and the best configuration of the grid otherwise (SINGC, a learning rate of $1/\nclust = 0.02$,
$\alpha = 0.1$, $\lambda = 0.1$), IPE returns three community-pure supernodes on all seeds,
for the $50$ requested (Fig.~\ref{fig:ipe-k}). Under the published defaults, the same rank still
returns $50$ supernodes with no community structure. Intuitively, the three leading eigenvectors of this SBM separate the
communities. On real graphs, where the classes are less tied to the spectrum, IPE may behave
differently.

\subsection{A random geometric graph}
\label{app:rgg}

We repeat the check of Sec.~\ref{sec:sbm} on a random geometric graph (RGG).
Its nodes are $500$ points drawn uniformly in the unit square, and two points closer than $0.08$ are joined. This gives a connected graph with $2306$ edges, i.e.\ a mean degree of about $9$.
The features are drawn as for the SBM, $X \sim \mathcal{N}(0, \Lap^\dagger)$. No method sees the positions.
Intuitively, a good coarsening of this graph forms supernodes that are compact in the square and connected in the graph, close to the Voronoi cells of their centroids.

\begin{figure}[H]
\centering
\includegraphics[width=0.95\linewidth]{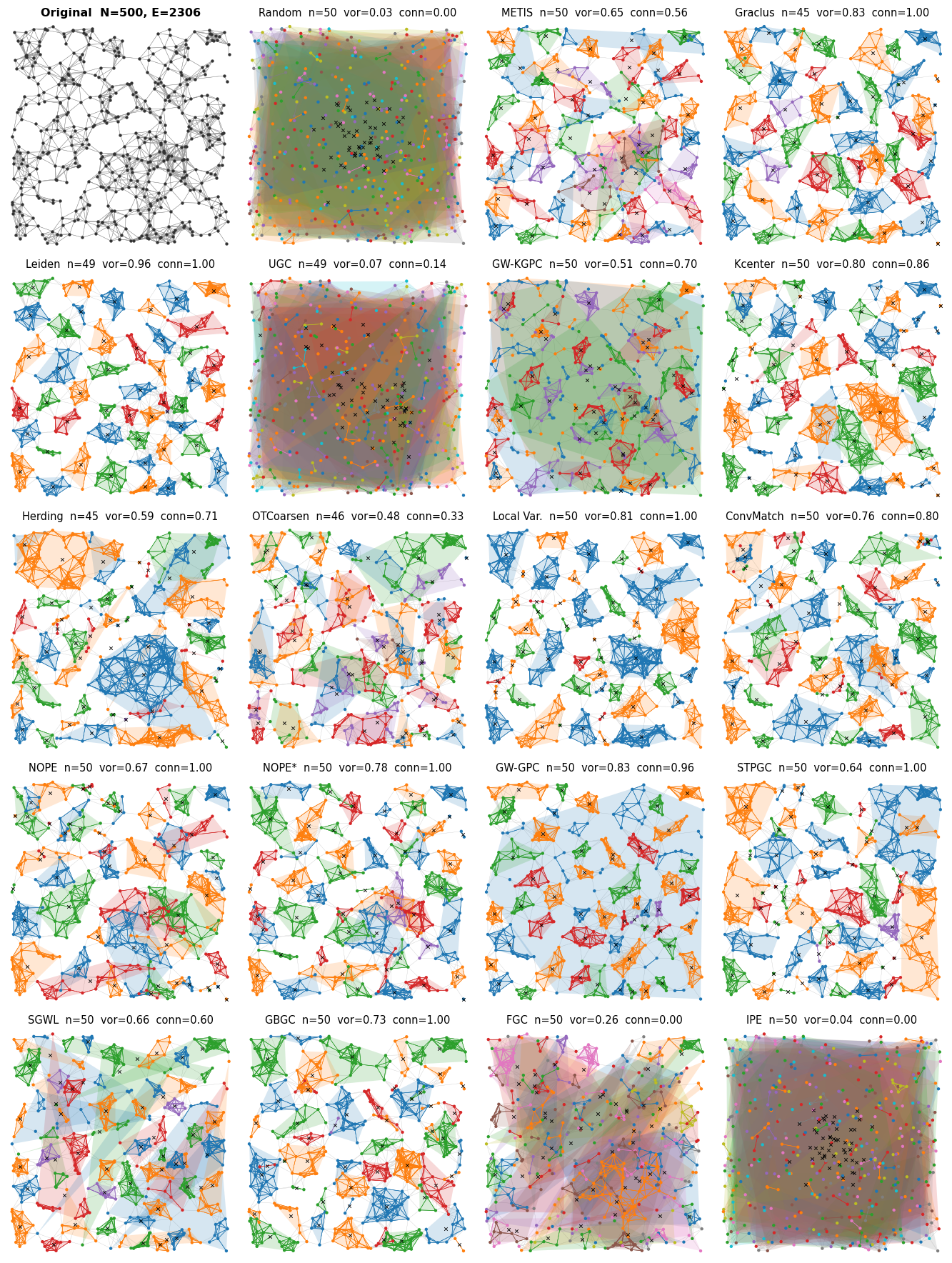}
\caption{Coarsened random geometric graph per method, at $r = 0.9$ ($50$ supernodes requested). The top-left panel is the original graph. Points are coloured by supernode, and two adjacent supernodes never share a colour. The shaded area is the convex hull of a supernode, and its inner edges are drawn in its colour. Crosses mark the centroids. A supernode that is not connected spreads its hull over other supernodes, and its centroid may fall outside its own points.}
\label{fig:rgg}
\end{figure}

We measure two quantities on the partition. The \textbf{Voronoi agreement} (vor) is the fraction of nodes whose nearest centroid, in the square, is the centroid of their own supernode. It equals $1$ when the partition is the Voronoi diagram of its centroids.
The \textbf{connectivity} (conn) is the fraction of supernodes whose nodes induce a connected subgraph of the original graph.
Every method runs at its default, except UGC at $\alpha = 0.5$, since its default computes $\alpha$ from the labels.
Fig.~\ref{fig:rgg} shows the coarsened graph of each method at $r = 0.9$. Table~\ref{tab:rgg} gives both quantities at $r = 0.7$ and $r = 0.9$.

Graclus, Leiden, local variation, NOPE, NOPE*, STPGC and GBGC return only connected supernodes at both ratios. This is expected from their construction. Graclus, local variation and NOPE only merge adjacent nodes or supernodes. STPGC merges a node into one of its neighbors. GBGC grows its balls by breadth-first search. Leiden guarantees connected communities~\cite{traag2019louvain}. It is also the closest to a Voronoi diagram, with $\text{vor} = 0.96$ at $r = 0.9$.
Random, UGC, FGC and IPE scatter their supernodes over the square, with $\text{vor} \leq 0.26$ and few or no connected supernodes at $r = 0.9$. Note that UGC, FGC and IPE run here without the tuning of App.~\ref{app:hp-sensitivity}.
METIS returns $56\%$ of connected supernodes at $r = 0.9$. METIS has an option that forces connected parts. It is off by default. Our wrapper keeps this default.
Several methods do not return exactly the requested number of supernodes. Table~\ref{tab:rgg} gives the number they return.

\begin{table}[H]
\centering
\small
\caption{Random geometric graph: number of supernodes returned ($n$), Voronoi agreement (vor) and connectivity (conn) of each method, for $150$ supernodes requested ($r = 0.7$) and $50$ ($r = 0.9$).}
\label{tab:rgg}
\begin{tabular}{lrrrrrr}
\toprule
& \multicolumn{3}{c}{$r = 0.7$} & \multicolumn{3}{c}{$r = 0.9$} \\
\cmidrule(lr){2-4}\cmidrule(lr){5-7}
Method & $n$ & vor & conn & $n$ & vor & conn \\
\midrule
Random          & 147 & 0.05 & 0.11 & 50 & 0.03 & 0.00 \\
METIS           & 150 & 0.46 & 0.72 & 50 & 0.65 & 0.56 \\
Graclus         & 145 & 0.72 & 1.00 & 45 & 0.83 & 1.00 \\
Leiden          & 150 & 0.92 & 1.00 & 49 & 0.96 & 1.00 \\
UGC             & 156 & 0.09 & 0.24 & 49 & 0.07 & 0.14 \\
GW-KGPC         & 150 & 0.65 & 0.83 & 50 & 0.51 & 0.70 \\
Kcenter         & 150 & 0.86 & 0.98 & 50 & 0.80 & 0.86 \\
Herding         & 142 & 0.78 & 0.92 & 45 & 0.59 & 0.71 \\
OTCoarsening    & 162 & 0.67 & 0.85 & 46 & 0.48 & 0.33 \\
Local variation & 150 & 0.94 & 1.00 & 50 & 0.81 & 1.00 \\
ConvMatch       & 150 & 0.84 & 0.94 & 50 & 0.76 & 0.80 \\
NOPE            & 150 & 0.62 & 1.00 & 50 & 0.67 & 1.00 \\
NOPE*           & 150 & 0.71 & 1.00 & 50 & 0.78 & 1.00 \\
GW-GPC          & 150 & 0.89 & 0.95 & 50 & 0.83 & 0.96 \\
STPGC           & 150 & 0.79 & 1.00 & 50 & 0.64 & 1.00 \\
SGWL            & 150 & 0.71 & 0.83 & 50 & 0.66 & 0.60 \\
GBGC            &  62 & 0.71 & 1.00 & 50 & 0.73 & 1.00 \\
FGC             & 147 & 0.21 & 0.40 & 50 & 0.26 & 0.00 \\
IPE             & 150 & 0.03 & 0.04 & 50 & 0.04 & 0.00 \\
\bottomrule
\end{tabular}
\end{table}

\section{Additional details}
\label{app:details}

\subsection{Storing and reloading a coarsening}
\label{app:storage}
A coarsening is stored as its partition. Each saved coarsening is a directory with \texttt{partition.npz} (the integer assignment and the
supernode count) and \texttt{meta.json} (the algorithm parameters, dataset, achieved
ratio, coarsening time, and peak memory). The full \texttt{Coarsening} is reconstructed on
load by replaying \texttt{build\_coarsening}. Storing the partition shrinks a saved coarsening compared with storing $\Adjc$, $\Lift$ and $\Proj$, and it can still be inspected with \texttt{np.load}. Local variation under its per-level reduction also saves its \texttt{P\_chain} in \texttt{partition.npz}.
Listing~\ref{lst:storage} shows how a coarsening is saved and reloaded. The reloaded \texttt{Coarsening} is identical to a freshly computed one.

\begin{lstlisting}[style=graco,
  caption={Storing and reloading a coarsening. Only the partition and its metadata are written to disk.},
  label={lst:storage}]
from graco import get_dataset, coarsen_local_variation, build_coarsening
from graco import save_coarsening, load_coarsening

data = get_dataset("Cora")
partition, info = coarsen_local_variation(data, r=0.5)
co = build_coarsening(data, partition, laplacian_kind="normalized_self_loop")

# save: only partition.npz and meta.json are written to disk
save_coarsening(co, dir_="runs/cora_lv_r05", dataset="Cora",
                algo="local_variation", algo_params={"r": 0.5})

# reload: the coarsened graph, Q and P are rebuilt by replaying build_coarsening
co = load_coarsening("runs/cora_lv_r05", data)
\end{lstlisting}

\section{Related surveys and tooling}
\label{app:related}

There are existing surveys on coarsening: Chen et al.~\cite{chen2022graph} review coarsening from
its scientific-computing origins, and Hashemi et al.~\cite{hashemi2024comprehensive} cover
it jointly with graph sparsification and condensation. These surveys organize the
literature, but none provides a codebase on which the methods can be run and compared.

The closest general-purpose tools address the neighboring object of graph pooling, which has
its own unifying abstractions and libraries: the Select-Reduce-Connect
framework~\cite{grattarola2022understanding} and Torch Geometric Pool~\cite{abate2025torch},
both built on PyTorch Geometric~\cite{fey2019fast} for pooling layers trained end-to-end
inside a graph neural network. A few of their non-trainable operators overlap with coarsening,
for instance Graclus~\cite{dhillon2007weighted}, but they cover neither the modern coarsening literature nor its
metrics. Pooling optimizes a task loss, while coarsening preserves a fixed graph property
(see Sec.~\ref{sec:notation}).

\end{document}